\documentclass{article}

\usepackage{microtype}
\usepackage{graphicx}
\usepackage{subcaption}
\usepackage{booktabs} 

\usepackage{hyperref}

\usepackage[accepted]{icml2026}

\usepackage{amsmath}
\usepackage{amssymb}
\usepackage{mathtools}
\usepackage{amsthm}

\usepackage{enumitem}
\usepackage{colortbl}
\usepackage{xcolor}
\usepackage{adjustbox}
\usepackage{multirow}
\definecolor{sectiongray}{gray}{0.92}
\usepackage[table]{xcolor}
\usepackage{makecell}
\usepackage{wrapfig} 
\usepackage{tcolorbox}
\tcbuselibrary{listings,breakable}
\newtcblisting{llmoutput}[1]{%
  breakable,
  colback=white,
  colframe=gray,
  coltitle=black,
  colbacktitle=gray!10,
  title=\textbf{#1},
  listing only,
  listing options={
    basicstyle=\ttfamily\small,
    breaklines=true,
    breakatwhitespace=true,
    columns=fullflexible,
    keepspaces=true
  }%
}

\usepackage[capitalize,noabbrev]{cleveref}

\theoremstyle{plain}
\newtheorem{theorem}{Theorem}[section]
\newtheorem{proposition}[theorem]{Proposition}

\theoremstyle{definition}

\theoremstyle{remark}

\usepackage[textsize=tiny]{todonotes}

\icmltitlerunning{Beyond Normalization: Rethinking the Partition Function as a Difficulty Scheduler for RLVR}

\begin{document}

\twocolumn[
  \icmltitle{Beyond Normalization: \\Rethinking the Partition Function as a Difficulty Scheduler for RLVR}


  \icmlsetsymbol{equal}{*}

  \begin{icmlauthorlist}
    \icmlauthor{Dohyung Kim}{snu}
    \icmlauthor{Minbeom Kim}{snu}
    \icmlauthor{Jeonghye Kim}{kaist}
    \icmlauthor{Sangmook Lee}{snu}
    \icmlauthor{Sojeong Rhee}{kaist}
    \icmlauthor{Kyomin Jung}{snu}
  \end{icmlauthorlist}

  \icmlaffiliation{snu}{Seoul National University}
  \icmlaffiliation{kaist}{KAIST}
  \icmlcorrespondingauthor{Dohyung Kim}{kimdohyung@snu.ac.kr}
  \icmlcorrespondingauthor{Kyomin Jung}{kjung@snu.ac.kr}

  \icmlkeywords{Machine Learning, ICML}

  \vskip 0.3in
]



\printAffiliationsAndNotice{}  
 
\begin{abstract}
Reward-maximizing RL methods have shown to be capable of enhancing the reasoning performance of LLMs, but often lead to reduced generation diversity. Recent works address this issue by adopting GFlowNets, training LLMs to match a target distribution while jointly learning its partition function. 
In contrast to prior works that treat this partition function solely as a normalizer, we reinterpret it as a per-prompt expected-reward (i.e., online accuracy) signal, leveraging this unused information to improve sample efficiency.
Specifically, we first establish a theoretical relationship between the partition function and per-prompt accuracy estimates. Building on this key insight, we propose \textbf{Pa}rtition Fun\textbf{c}tion-Guid\textbf{ed} \textbf{RL} (PACED-RL), a  post-training framework that leverages accuracy estimates to prioritize informative question prompts during training, and further improves sample efficiency through an accuracy estimate error–prioritized replay. Crucially, both components reuse information already produced during GFlowNet training, effectively amortizing the compute overhead into the existing optimization process. Extensive experiments across diverse benchmarks demonstrate strong performance improvements over GRPO and prior GFlowNet approaches, highlighting PACED-RL as a promising direction for a more sample efficient distribution-matching training for LLMs.
\end{abstract}

\section{Introduction}
Reinforcement Learning (RL) has emerged as a cornerstone for improving the reasoning performance of Large Language Models (LLMs) \citep{jaech2024openai,guo2025deepseek,comanici2025gemini}. By enabling models to repeatedly self-explore and maximize the expected reward, such methods have been shown to significantly enhance complex reasoning performances \citep{shao2024deepseekmath,yu2025dapo}.

Despite these gains, \textit{reward-maximizing} RL methods such as PPO \citep{schulman2017proximal} and GRPO \citep{shao2024deepseekmath} often lead to an overly sharpened output distribution \citep{huang2025selfimprovement,li2025preserving}, leading to limited diversity among generated outputs \citep{padmakumar2024does,shypula2025evaluating}. In the context of RL post-training for LLM reasoning, however, preserving diversity is essential for a thorough exploration of the search space, enabling the discovery of diverse valid reasoning strategies and the effective application of inference-time techniques \cite{yue2025does,li2025jointly,chen2025pass}.

To alleviate this issue, recent methods such as FlowRL \citep{zhu2026flowrl} depart from direct reward maximization and instead employ GFlowNets \citep{bengio2023gflownet}, focusing on reward \textit{distribution matching}. Unlike post-training with reward-maximizing RL, where the LLM policy is optimized to maximize the expected reward, FlowRL trains the policy to match a reward-induced target distribution, with a partition function jointly learned as the normalizer of that target reward distribution. At optimum, this partition function corresponds to the sum of the reward-induced distribution mass, accumulated over all possible completions for a given input question prompt. By explicitly modeling the target distribution and pushing the policy toward it, this distributional objective encourages mode coverage, thereby improving upon the diversity limitations of conventional RL methods.

In this work, we revisit the GFlowNet partition function used in LLM post-training. We show that the partition function---previously regarded only as a necessary overhead for target distribution normalization and training stability---naturally encodes per-prompt online accuracy information that can be leveraged for significant improvements in sample efficiency. Specifically, we establish a theoretical connection between the partition function and online accuracies, showing that it can be used directly to estimate the accuracy of a given question prompt under the policy during training. This insight allows us to re-purpose the cost already incurred by GFlowNet training to guide adaptive prompt selection and replay prioritization, focusing on the most informative training samples at each step.

Building on this observation, we propose \textbf{Pa}rtition Fun\textbf{c}tion-Guid\textbf{ed} \textbf{RL} (PACED-RL), a novel GFlowNet-based LLM post-training framework that extends the role of the partition function beyond normalization to enable a more sample efficient training, while preserving the inherent diversity benefits of GFlowNets. By making use of the partition function to directly predict accuracy estimates, our method performs difficulty-aware adaptive prompt selection, sampling question prompts that maximize learning efficiency at each step. Furthermore, inspired by prior work on replay mechanisms \citep{schaul2015prioritized,shen2023towards,kim2024local}, we introduce an accuracy estimation error–prioritized replay strategy that exploits the off-policy tolerance of the GFlowNet objective to further boost sample efficiency. Crucially, both components reuse information \emph{already} produced during standard GFlowNet training, effectively amortizing the cost of adaptive prompt selection and replay prioritization into the existing optimization process.

Extensive experiments across mathematical reasoning and coding benchmarks showcase the effectiveness of PACED-RL. On AIME benchmarks, PACED-RL improves average pass@1 performance by up to 29.1\% and 40.0\% over GRPO and FlowRL, respectively. Moreover, on the pass@$k$ metric, which we employ as a proxy for  diversity and exploration capacity \citep{li2025jointly,zhu2025the}, PACED-RL achieves consistent improvements over baselines, outperforming GRPO and FlowRL by up to 14.2\% and 9.1\%.

Our main results and key contributions are as follows:
\begin{itemize}[leftmargin=*,nosep]
    \item We theoretically show that the GFlowNet partition function for LLM post-training is linked to online accuracies, enabling it to directly serve as online accuracy estimates.
    \item We introduce PACED-RL, which leverages these accuracy estimates to perform difficulty-aware adaptive prompt selection and estimation-error–prioritized replay, boosting training efficiency while preserving output diversity.
    \item We carry out extensive evaluations across code and mathematical reasoning tasks, and show that PACED-RL achieves clear performance gains over recent baselines. 
\end{itemize}

Overall, our results show that PACED-RL offers a principled and practical way to further enhance distribution-matching training of LLMs for reasoning tasks.
\section{Preliminaries}
\subsection{Reinforcement Learning with Verifiable Rewards}
Reinforcement Learning with Verifiable Rewards (RLVR) has emerged as an effective approach for training LLMs on verifiable tasks. Given an output $\mathbf{y}$ generated by an LLM policy $\pi_{\theta}$ for an input question prompt $\mathbf{x}$ sampled from a dataset $\mathcal{D}$, a deterministic reward function $r(\mathbf{x}, \mathbf{y}) \in \{0, 1\}$ assigns a binary score to $\mathbf{y}$ by assessing its correctness. RLVR optimizes the following KL-regularized objective:
\begin{equation}
\label{eq:rlvr_objective}
\underset{\theta}{\max}\,
\!\!\mathop{\mathbb{E}}\limits_{\substack{\mathbf{x} \sim \mathcal{D} \\ \mathbf{y} \sim \pi_{\theta}(\cdot \mid \mathbf{x})}}
\!\!\!\left[r(\mathbf{x}, \mathbf{y})\right]
-
\beta D_{\mathrm{KL}}\!\left(\pi_{\theta}(\cdot\! \mid \mathbf{x}) \| \pi_{\mathrm{ref}}(\cdot\! \mid \mathbf{x})\right)
\end{equation}
, where $\pi_{\mathrm{ref}}$ denotes the untrained base reference model and $\beta$ denotes the KL divergence regularization coefficient.

In particular, the optimal policy for Eq. \ref{eq:rlvr_objective} admits a closed-form solution, expressed as:
\begin{equation}
\label{eq:optimal_policy}
\pi^{*}(\mathbf{y} \mid \mathbf{x})
\;=\;
\frac{
    \pi_{\mathrm{ref}}(\mathbf{y} \mid \mathbf{x})\,
    \exp\!\left(\beta^{-1} r(\mathbf{x}, \mathbf{y})\right)
}{
    Z(\mathbf{x})
}.
\end{equation}
Here, $Z(\mathbf{x})$ denotes the intractable partition function that normalizes the distribution  by summing over all possible outputs $\mathbf{y}$ for a given question prompt $\mathbf{x}$:
\begin{equation}
\label{eq:true_partition}
Z(\mathbf{x}) = \sum_{\mathbf{y}} \pi_{\mathrm{ref}}(\mathbf{y} \mid \mathbf{x})\,\exp\!\left(\beta^{-1} r(\mathbf{x}, \mathbf{y})\right).
\end{equation}
\subsection{GFlowNets for LLM Post-Training}
GFlowNets train a policy $\pi_{\theta}$ to sample diverse discrete, compositional objects in proportion to an unnormalized target reward \textit{distribution} $R(\mathbf{x}, \mathbf{y})$. In the RLVR setting, this objective biases the policy to sample high-reward outputs $\mathbf{y}$ (correct solutions) rather than low-reward outputs (incorrect solutions) for a given input question prompt $\mathbf{x}$.

To train an LLM policy $\pi_{\theta}$ to sample from the optimal policy in Eq. \ref{eq:optimal_policy} using GFlowNet objectives, previous works \citep{lee2025learning,zhu2026flowrl} configure the unnormalized reward distribution as $R(\mathbf{x},\mathbf{y}) = \pi_{\mathrm{ref}}(\mathbf{y} \mid \mathbf{x}) \exp\!\left(\beta^{-1} r(\mathbf{x},\mathbf{y})\right)$, and train a learnable $Z_{\phi}(\mathbf{x})$, which approximates the intractable partition function $Z(\mathbf{x})$ in Eq. \ref{eq:true_partition}. By explicitly modeling the target distribution and its normalization through a learnable partition function, GFlowNets reduce policy learning to matching this target distribution.

Plugging this $R(\mathbf{x},\mathbf{y})$ and $Z_{\phi}(\mathbf{x})$ into the Trajectory Balance (TB) objective \citep{malkin2022trajectory} for GFlowNet training leads to the loss function defined as follows:
\begin{equation}
\label{eq:tb_loss}
\mathcal{L}_{\mathrm{TB}}(\mathbf{x},\mathbf{y};\theta,\phi)
\!=\!\!
\left[
    \log\!\left(
        \frac{
            Z_{\phi}(\mathbf{x})\, \pi_{\theta}(\mathbf{y} \mid \mathbf{x})
        }{
            \pi_{\mathrm{ref}}(\mathbf{y} \mid \mathbf{x})\,
            \exp\!\bigl(\beta^{-1} r(\mathbf{x}, \mathbf{y})\bigr)
        }
    \right)
\right]^{2}\!\!.
\end{equation}
Minimizing this TB loss is equivalent, in terms of expected gradients, to minimizing the KL divergence between $\pi_{\theta}$ and the optimal policy in Eq. \ref{eq:optimal_policy}, with the intractable partition function approximated by $Z_{\phi}(\mathbf{x})$ \citep{zhu2026flowrl}: 
\begin{equation}
\label{eq:kl_is_tb}
\begin{aligned}
&\min_{\theta}\;
\mathcal{L}_{\mathrm{TB}}(\mathbf{x},\mathbf{y};\theta,\phi)
\;\Longleftrightarrow\;
\\
&\min_{\theta}\;
D_{\mathrm{KL}}\!\left(
\pi_{\theta}(\mathbf{y}\mid\mathbf{x})
\;\middle\|\;
\frac{
    \pi_{\mathrm{ref}}(\mathbf{y} \mid \mathbf{x})\,
    \exp\!\left(\beta^{-1} r(\mathbf{x}, \mathbf{y})\right)
}{
    Z_{\phi}(\mathbf{x})
}
\right).
\end{aligned}
\end{equation}
Thus, $\mathcal{L}_{\mathrm{TB}}$ drives the LLM policy $\pi_{\theta}$ to sample from the optimal policy, while simultaneously training $Z_{\phi}(\mathbf{x})$ to normalize the unnormalized target reward distribution $\pi_{\mathrm{ref}}(\mathbf{y} \mid \mathbf{x}) \exp\!\left(\beta^{-1} r(\mathbf{x}, \mathbf{y})\right)$. 

\section{Related Works}
\subsection{GFlowNets for LLM Post-Training}
 In the context of LLM post-training, GFlowNets have been adapted for domains such as preference alignment, red-teaming, and reasoning \citep{kwon2024gdpo, lee2025learning, yu2025flow,zhu2026flowrl}. \citet{bartoldson2025trajectory} further proposes an asynchronous, distributed RL pipeline for LLMs that leverages GFlowNets for better throughput. 

Despite their success, existing GFlowNet-based methods for LLM post-training
primarily treat the partition function as a necessary normalization variable
required to define the target distribution.
While several works study improved optimization techniques for GFlowNets, such as replacing the learned $Z_{\phi}(\mathbf{x})$ with a batch-estimate \citep{zhang2023robust,bartoldson2025trajectory}, the information encoded in $Z_{\phi}(\mathbf{x})$ has not been exploited beyond normalization.

Most closely related to our work is FlowRL \citep{zhu2026flowrl}. FlowRL adapts GFlowNets to the synchronous RL setting, where training alternates between the rollout generation phase and the optimization phase, with added stability techniques suited for RLVR. Our work extends prior works in GFlowNets for LLM post-training by reformulating $Z_{\phi}(\mathbf{x})$ for online accuracy estimation, enabling adaptive prompt selection and replay strategies that enhance sample efficiency without incurring additional cost. 

\subsection{Adaptive Prompt Selection for RLVR}
Recent works show that training on question prompts of intermediate difficulty—those achieving approximately 0.5 accuracy with respect to the policy—improves sample efficiency in the RLVR setting \citep{bae-etal-2026-online,foster2026lilo}. Building on this observation, adaptive prompt selection methods that selectively sample such prompts for training have emerged as a promising direction for improved sample efficiency. Crucially, to effectively enable such methods, obtaining reliable online accuracy estimates is needed.

Recent works such as \citet{yu2025dapo,foster2026lilo,zhang2025learning} over-sample a pool of question prompts larger than the training batch at each step, and filter out less informative prompts based on the observed accuracies after the rollout generation phase. Though effective, such approaches lead to a significantly increased number of rollout generation, substantially increasing computational overhead. An alternative line of work maintains per-question accuracy histories, and estimates online accuracies via Bayesian posterior estimation or probabilistic filtering \citep{zheng2025act,qu2026can,zeng2026cures}, enabling lightweight prompt selection. However, on large datasets, these estimates can suffer from off-policy bias: over the course of an epoch, the policy may improve substantially, rendering the accuracy estimates stale when revisited. 

Distinct from previous approaches, our work reuses the partition function inherent to GFlowNets for online accuracy estimation, eliminating the need for employing such auxiliary mechanisms to guide adaptive prompt selection.

\subsection{Prioritized Experience Replay}
Replay is widely adopted in RL training, owing to its efficacy in enhancing sample efficiency. To further enhance replay effectiveness, \citet{schaul2015prioritized} propose Prioritized Experience Replay (PER) for Q-learning, which prioritizes samples with large temporal-difference errors, allowing the policy to focus learning on more informative data. Building on this idea, numerous PER variants \citep{horgan2018distributed,hessel2018rainbow,sujit2023prioritizing} have been developed to improve training stability and data efficiency. 

Replay mechanisms have also been extensively adopted in the GFlowNets literature \citep{shen2023towards,kim2024local,bartoldson2025trajectory}, where the training objective naturally accommodates off-policy data, and more recently, have been explored in the RLVR setting \citep{wang2025eframe,li2025repo}. A recurring insight across both domains is that prioritizing high-reward and more recent samples improves the effectiveness of the replay. Extending upon previous works, our work tailors prioritization to improve partition function learning and sample efficiency by leveraging its connection to accuracy estimates and prioritizing samples with large accuracy estimation errors.
\section{PACED-RL}
We begin by revisiting the role of the GFlowNet partition function for LLM post-training. While the learnable partition function $Z_{\phi}$ is typically introduced as a necessary normalization term, we reveal that it in fact encodes meaningful information about per-question online accuracies. Specifically, we first show that for a question $\mathbf{x}$, $Z_{\phi}$ provides an estimate of the accuracy $p_{\mathrm{old}}(\mathbf{x})$. Here, $p_{\mathrm{old}}(\mathbf{x})$ denotes the accuracy for $\mathbf{x}$ under the pre-update policy $\pi_{\mathrm{old}}$, the policy in which fresh rollouts are sampled from at each training step. We then show how these accuracy estimates can be leveraged to improve sample efficiency by focusing on the most informative samples throughout the training process.
\subsection{Partition Function and Online Accuracy Estimates}
\label{sec:partition}
Under a Trajectory Balance loss modified from Eq. \ref{eq:tb_loss}, we show that the optimal partition function admits a direct relationship to per-question online accuracy estimates.
Specifically, we replace $\pi_{\mathrm{ref}}$ with $\pi_{\mathrm{old}}$:
\begin{equation}
\label{eq:our_tb_loss}
\mathcal{L}_{\mathrm{ours}}(\mathbf{x},\mathbf{y};\theta,\phi)
\!=\!
\left[
    \log\!\left(\!\!
        \frac{
            Z_{\phi}(\mathbf{x})\, \pi_{\theta}(\mathbf{y} \mid \mathbf{x})
        }{
            \pi_{\mathrm{old}}(\mathbf{y} \mid \mathbf{x})\,
            \exp\!\bigl(\beta^{-1} r(\mathbf{x}, \mathbf{y})\bigr)
        }
    \!\!\right)
\!\right]^{2}\!\!\!.
\end{equation}
Replacing the reference policy $\pi_{\mathrm{ref}}$ with $\pi_{\mathrm{old}}$ yields an interpretation as training to sample from the optimal policy of a KL-regularized objective (Eq.~\ref{eq:rlvr_objective}), with the KL divergence regularization anchored at $\pi_{\mathrm{old}}$ rather than $\pi_{\mathrm{ref}}$.
\begin{proposition}
    \label{prop:main}
    Given $Z^*(\mathbf{x})$, the optimal partition function, we can express $p_{\mathrm{old}}(\mathbf{x})$ as follows:
    \begin{equation}
    \label{eq:acc}
        p_{\mathrm{old}}(\mathbf{x})
        \!=
        \!\beta \log Z^{*}(\mathbf{x})
        -
        \beta D_{\mathrm{KL}}\!\left(
        \pi_{\mathrm{old}}(\cdot \!\mid\! \mathbf{x})
        \,\!\big\|\, \!
        \pi_{\theta}(\cdot\! \mid\! \mathbf{x})
        \right).
    \end{equation}
\end{proposition}
The detailed derivations are provided in Appendix \ref{app:proofs}. Proposition \ref{prop:main} reveals that the partition function, previously used solely for the normalization of the target distribution, can be directly interpreted as an online accuracy signal. Importantly, this information is already produced as part of the GFlowNet training, incurring no extra computational cost.

\begin{figure}[ht]
  \centerline{\includegraphics[width=0.7\columnwidth]{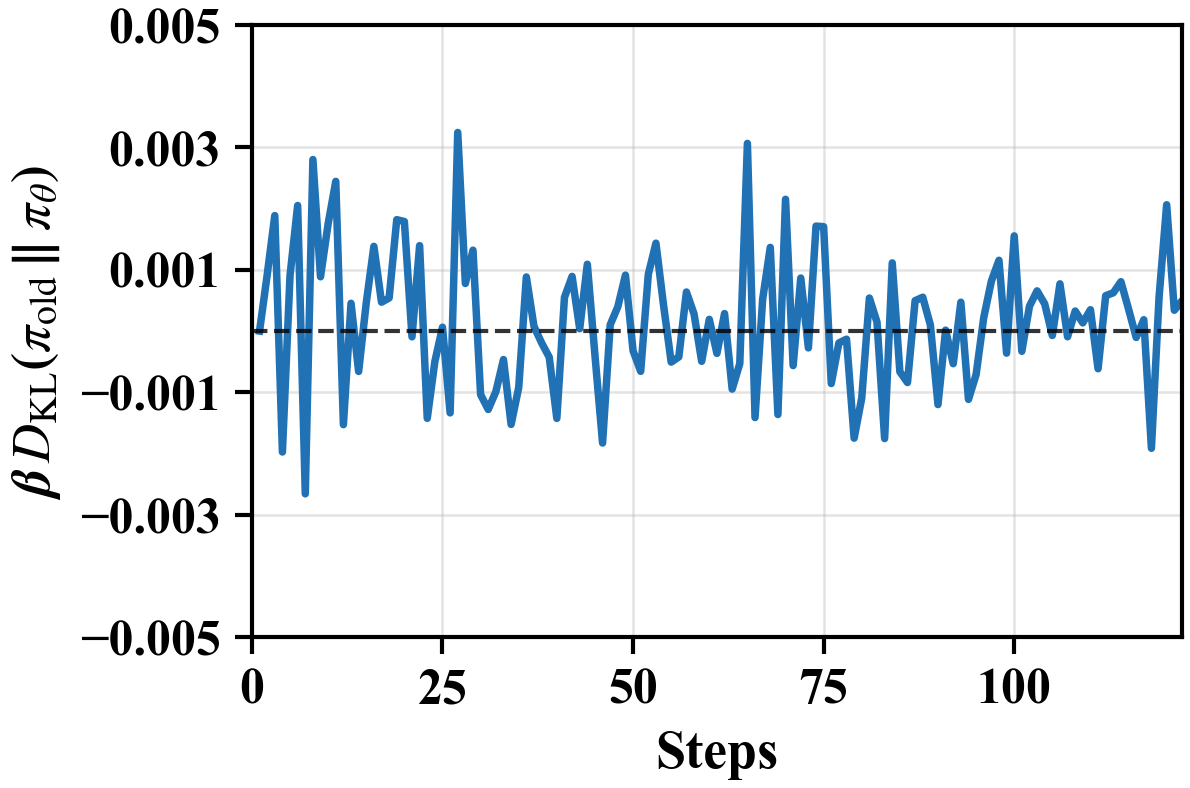}}
    \caption{
      Training dynamics of mean of estimated values of  $
\beta\,D_{\mathrm{KL}}\!\left(\pi_{\mathrm{old}}(\cdot\!\mid \mathbf{x}) \,\!\big\|\!\, \pi_{\theta}(\cdot\!\mid \mathbf{x})\right)
$ of Qwen2.5-Math-1.5B trained on the DeepScaleR dataset, using the TB loss defined in Eq. \ref{eq:our_tb_loss}}
    \label{fig:residual}
\end{figure}

\subsection{Obtaining Practical Online Accuracy Estimates}
 Ideally, the accuracy estimate for $\mathbf{x}$ should be obtainable as a function of $\mathbf{x}$ without requiring additional computation. While Eq. \ref{eq:acc} provides theoretical insight, it does not immediately yield a practical online accuracy estimate from $Z_{\phi}$, as KL divergence between policies is typically estimated using expensive Monte Carlo rollouts \citep{schulman2015trust}. Below, we show that the KL term in Eq. \ref{eq:acc} remains uniformly small during training and can therefore be safely omitted, yielding a simple accuracy estimator dependent only on $Z_{\phi}$.

In practice, standard training practices—such as small learning rates and gradient norm clipping—generally ensure that parameter updates remain small. As KL divergence admits a second-order approximation that scales quadratically with the parameter update \citep{achiam2017constrained,martens2020new}, small parameter updates typically lead to a controlled KL divergence between successive policies. 

As shown in Fig.~\ref{fig:residual}, the empirically estimated mean values of $
\beta\,D_{\mathrm{KL}}\!\left(\pi_{\mathrm{old}}(\cdot\!\mid \mathbf{x}) \,\!\big\|\!\, \pi_{\theta}(\cdot\!\mid \mathbf{x})\right)
$ remains uniformly small over the course of training, with the maximum absolute value remaining below $4\times10^{-3}$.
Since the range of online accuracies $p_{\mathrm{old}}(\mathbf{x})$ is $[0,1]$ due to the 0/1  binary reward setting, the error that would be introduced by discarding the KL divergence term is small at the scale relevant for difficulty-aware prompt selection. 

Motivated by this observation, we obtain a practical biased estimator by discarding the KL divergence term in Eq.~\ref{eq:acc}:
\begin{equation}
    \label{eq:accuracy_final}
    p_{\mathrm{old}}(\mathbf{x})
    \approx
    \beta \log Z^{*}(\mathbf{x})
\end{equation}
and train $Z_{\phi}(\mathbf{x})$, which approximates the optimal $Z^{*}(\mathbf{x})$. While this approximation introduces an empirically stable and controlled bias into the accuracy estimates, it enables per-question online accuracy estimation without auxiliary estimators or additional rollouts. Consequently, the cost of accuracy estimation is effectively amortized into the existing GFlowNet training process.

\subsection{Adaptive Prompt Selection Using Online Accuracy Estimates}
Having obtained online accuracy estimates directly from $Z_{\phi}$ \textit{without} incurring additional cost, we can naturally integrate adaptive prompt selection into the training pipeline. 

Prior work has shown that question prompts of intermediate difficulty provide the most sample-efficient learning signal \citep{bae-etal-2026-online,foster2026lilo}. Leveraging the online accuracy estimates from $Z_{\phi}$, we therefore bias training toward question prompts whose predicted accuracies lie in the vicinity of $0.5$, ensuring that the policy consistently focuses on the most informative prompts during training.

Specifically, at training step $t$, we begin by utilizing Eq. \ref{eq:accuracy_final} to obtain $\{\hat{p}_{\mathrm{old}}(\mathbf{x}_i)\}_{i=1}^{|\mathcal{D}|}$, the estimates of the online accuracies for question prompts in the dataset. Similar to previous works \citep{yue2025does,foster2026lilo}, we then greedily select the top-$m$ questions whose estimated accuracies are closest to the target accuracy $\tau$, where $m$ denotes the training batch size. The selected $m$ questions are then used for the $t$-th training step. While we fix $\tau$ to $0.5$ for optimal sample efficiency in the main experiments, we report results for other configurations of $\tau$ in Sec.~\ref{sec:analysis}.

Following the FlowRL pipeline, for each of the $m$ selected questions, we then generate $N$ rollouts and update both the policy $\pi_{\theta}$ and the partition function $Z_{\phi}$ by minimizing Eq.~\ref{eq:our_tb_loss}.

\subsection{Accuracy Estimation Error-Prioritized Replay} 
To further improve sample efficiency and the calibration of $Z_{\phi}$, we introduce an accuracy estimation error–prioritized replay strategy. Unlike prior replay prioritization approaches \citep{schaul2015prioritized,shen2023towards}, our method exploits the connection between $Z_{\phi}$ and online accuracy estimates by focusing on samples with large estimation errors. Intuitively, these samples provide the most informative learning signal for $Z_{\phi}$; this leads to better-calibrated accuracy estimates and a more accurate normalization of the target reward distribution, thereby improving training.

Specifically, we maintain a replay buffer that stores prompt–output pairs $\{\mathbf{x}, \mathbf{y}\}$. Following prior findings that replay is most effective with high-reward and recent samples \citep{bartoldson2025trajectory,li2025repo}, we restrict the buffer to outputs satisfying $r(\mathbf{x},\mathbf{y})=1$. Among retained pairs, we prioritize pairs associated with question prompts $\mathbf{x}$ with the highest accuracy estimation error:
\begin{equation}
\label{eq:priority}
\mathrm{priority}(\mathbf{x})
=
\left|
\frac{N_{\mathrm{correct}}}{N}
-
\hat{p}_{\mathrm{old}}(\mathbf{x})
\right|,
\end{equation}
, where $\frac{N_{\mathrm{correct}}}{N}$ denotes the observed accuracy for question $\mathbf{x}$ computed from $N$ rollouts generated by $\pi_{\mathrm{old}}$ during the rollout generation process of the training pipeline.

We utilize a  fixed-capacity replay buffer of size $B_{\mathrm{max}}$ into which $B_{\mathrm{add}}$ new samples are added at every training step. This design bounds the age of stored trajectories and thereby controls the degree of off-policyness of the buffer \citep{fedus2020revisiting}. We then augment the training batch with the contents stored in the replay buffer at each training step. 

A complete description of our full PACED-RL algorithm, consisting of adaptive prompt selection and accuracy estimate error-prioritized replay, is shown in Algorithm \ref{alg:our_method}.
\section{Experiments}
\subsection{Experimental Setup}
\paragraph{Datasets \& Models.}
We conduct experiments in the code generation domain and the mathematical reasoning domain. For code generation, we train on the DeepCoder dataset \citep{luo2025deepcoder,zhu2026flowrl} using the DeepSeek-R1-Distill-Qwen-1.5B model \citep{guo2025deepseek}. For mathematical reasoning, we use the DeepScaleR dataset \citep{luo2025deepscaler} and adopt Qwen2.5-Math-1.5B and Qwen2.5-Math-7B \citep{yang2024qwen2} as base models. Across all runs, PACED-RL is trained with $\beta=0.05$ and replay buffer capacity $B_{\mathrm{max}}$ of 128, with $B_{\mathrm{add}}=64$ prompt-output pairs added to the buffer at each step. Similar to prior works \citep{lee2025learning,zhu2026flowrl}, we use a 3-layer MLP stacked on top of frozen, pre-computed reference model embeddings of question prompts $\mathbf{x}$ to parameterize $Z_{\phi}$.
\paragraph{Training \& Evaluation.}
We adopt the widely used verl
library \citep{sheng2025hybridflow} 
for training. Across all runs,
we fix the learning rate to 1e-6 and generate 8 rollouts per question prompt, i.e. $N=8$. Due to compute resource constraints, we set the maximum generation length to 3072 tokens for all runs. For code generation, we evaluate on HumanEval+ \citep{chen2021evaluating} and LiveCodeBench \citep{jain2025livecodebench}, and report the pass@1 performance averaged over 8 attempts throughout training. For evaluation on the mathematical reasoning domain,
we use the MATH500 \citep{hendrycks2021measuring}, MinervaMath
\citep{lewkowycz2022solving}, OlympiadBench \citep{he2024olympiadbench}, and AIME24/25 benchmarks. Similar to prior works \citep{wang2025eframe, zheng2025act}, we evaluate models every 10 steps and report the pass@1 and pass@$k$ performance of
the checkpoint with the best average performance across
benchmarks. Hyperparameters
and other implementational details are in Appendix \ref{app:experimental_details}.
\paragraph{Baselines.}
To comprehensively evaluate PACED-RL, we compare against representative baselines. 
\\\textbf{(1) No Adaptive Prompt Selection:}
We include \textbf{GRPO} and \textbf{FlowRL} as standard baselines that sample prompts uniformly from the training dataset. Neither method performs adaptive prompt selection; \textbf{GRPO} represents the reward-maximization approach, while \textbf{FlowRL} represents the distribution-matching algorithm based on GFlowNets.
\\\textbf{(2) With Adaptive Prompt Selection:}
We include Dynamic Sampling (\textbf{DS}) \citep{yu2025dapo} and \textbf{LILO} \citep{foster2026lilo} as baselines that perform adaptive prompt selection by over-sampling $M>m$ question prompts from the dataset at each training step. After generating rollouts for each of the $M$ over-sampled prompts, DS discards question prompts with observed accuracies of 0 or 1, whereas LILO selects the top-$m$ question prompts whose accuracies are closest to 0.5. We also compare PACED-RL against \textbf{MoPPS} \citep{qu2026can}, which estimate prompt accuracies without additional over-sampling. MoPPS maintains per-question prompt accuracy histories and applies Bayesian posterior estimation, selecting the top-$m$ question prompts whose estimated accuracies are closest to 0.5 at each training step.
\begin{figure}[ht]
    \centerline{\includegraphics[width=\columnwidth]{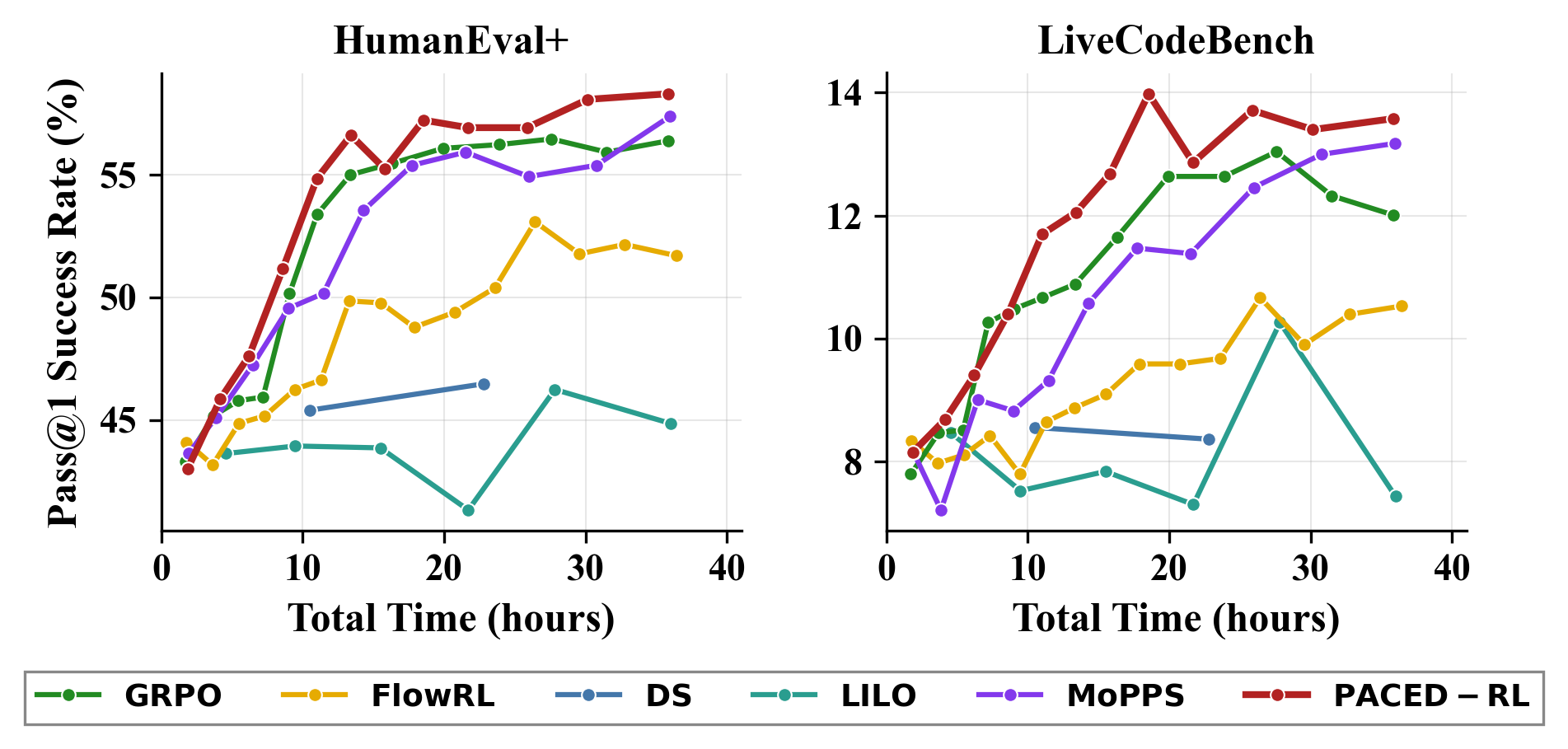}}
    \caption{
      Pass@1(\%) dynamics of DeepSeek-R1-Distill-Qwen-1.5B plotted with respect to wall-clock time. Pass@1 values obtained via averaging over 8 independent attempts. 
    }
    \label{fig:coding}
\end{figure}
\begin{figure}[t]
    \centerline{\includegraphics[width=\columnwidth]{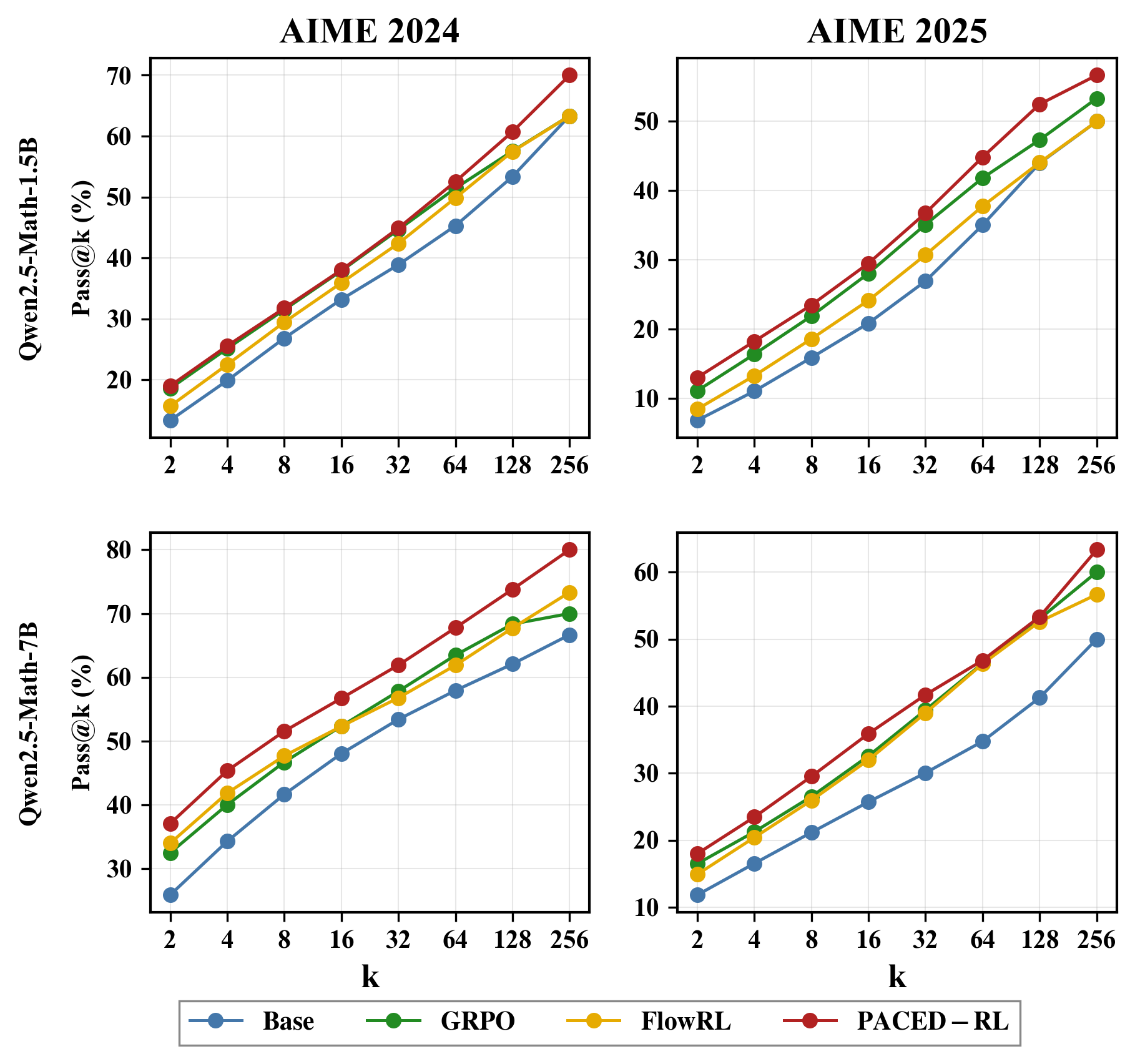}}
    \caption{
      Pass@k(\%) curves of Qwen2.5-Math-7B trained on GRPO, FlowRL, and PACED-RL and evaluated on AIME 24 and AIME24 for $k \in \{2,4,8,16,32,64,128,256\}$. Across all evaluated values of $k$, PACED-RL achieves the highest performance.
    }
    \label{fig:passk}
\end{figure}
\subsection{Main Results}
\paragraph{Code Generation.}
As shown in Fig. \ref{fig:coding}, PACED-RL consistently achieves the highest pass@1 performance throughout training compared to all baselines, illustrating the benefits of improved sample efficiency via leveraging accuracy estimates from the partition function. Notably, PACED-RL achieves significantly faster training than GRPO and FlowRL. On HumanEval+, PACED-RL reaches the best performance achieved by GRPO and FlowRL using only 0.49$\times$ and 0.42$\times$ of their respective training time. Similarly, on LiveCodeBench, PACED-RL achieves the best performance of GRPO and FlowRL in 0.67$\times$ and 0.42$\times$ of the time.

Our results also indicate that for code generation tasks, adaptive prompt selection methods that utilize over-sampling question prompts, namely DS and LILO, are particularly ineffective. Unlike the mathematical reasoning domain, where fast rule-based verification is possible, the code generation domain requires slow execution-based verification \citep{wang2025nemotron}. The verification of increased number of code generations to estimate accuracies required in the DS and the LILO settings leads to prohibitively slow training --- leading to sub-optimal results with respect to training latencies.

\paragraph{Mathematical Reasoning.} 
\begin{table*}[t]
\caption{Pass@1(\%) performance across the five mathematical reasoning benchmarks. We report the Avg@32 performance for AIME 24 and AIME 25 benchmarks due to their small size. \textbf{Avg.} denotes the average of the performance on each of the benchmarks. \textbf{Avg. Rollouts} denotes the average number of rollouts generated per training step relative to GRPO, expressed as a multiplicative factor. Best results are in \textbf{bold}. Second best results are \underline{underlined}.}
\centering
\footnotesize
\begin{adjustbox}{max width=\textwidth,center}
\begin{tabular}{llccccc|cc}
\toprule
\textbf{Model}
& \textbf{Method} 
& \textbf{MATH-500}
& \textbf{OlympiadBench}
& \textbf{Minerva} 
& \makecell{\textbf{AIME 24}\\\textbf{Avg@32}} 
& \makecell{\textbf{AIME 25}\\\textbf{Avg@32}} 
& \textbf{Avg.\,\textuparrow}
& \textbf{Avg. Rollouts\,\textdownarrow}
 \\
\midrule

\multirow[c]{6}{*}{\centering Qwen2.5-Math-1.5B}
& GRPO   &  70.6& 34.2 & 24.2 &10.3 & 6.3&29.1&\textbf{1.0}$\times$ \\
& FlowRL &  67.6& 32.4 & 20.2 & 9.5 &  5.9&27.1&\textbf{1.0}$\times$\\
\cmidrule(lr){2-9}
& DS   & \underline{72.8} & \underline{36.8} &  \underline{28.3}&11.8& 6.5  & \underline{31.2}&2.0$\times$\\
& LILO   & \textbf{73.2} & 36.2 & 27.9 &\underline{12.0}&  \underline{6.9}&\underline{31.2}&4.0$\times$\\
& MoPPS   & 71.6 & 36.6 & 26.1 & 11.4 & 6.6   & 30.4&\textbf{1.0}$\times$\\
\rowcolor{gray!15}
\cellcolor{white}{} 
& PACED-RL   & \textbf{73.2} & \textbf{37.9} & \textbf{29.0}  & \textbf{13.3}&  \textbf{7.3}&\textbf{32.1}&\textbf{1.0}$\times$\\
\midrule

\multirow[c]{6}{*}{\centering Qwen2.5-Math-7B}
& GRPO   &77.6  &39.7  & \underline{34.9}  &23.9  & 10.3  &37.2&\textbf{1.0}$\times$ \\
& FlowRL & 76.4 & 39.6 & 32.3  &24.9  & 9.4&36.5 &\textbf{1.0}$\times$\\
\cmidrule(lr){2-9}
& DS   & \textbf{80.2} & 44.0 &34.5  & 26.6 &12.0 &\underline{39.4}&2.5$\times$\\
& LILO   &78.4  &\textbf{45.2} & \textbf{37.5}  &\underline{28.0}  & 11.7& \textbf{40.1}&4.0$\times$\\
& MoPPS   & \textbf{80.2} & 42.2 & 34.1  & 24.0& \underline{12.9} &38.6&\textbf{1.0}$\times$\\
\rowcolor{gray!15}
\cellcolor{white}{} 
& PACED-RL   & \underline{80.0} & \underline{44.6} &34.1 & \textbf{28.7} & \textbf{13.1}   &\textbf{40.1}&\textbf{1.0}$\times$\\
\bottomrule
\end{tabular}
\end{adjustbox}

\vspace{3pt}
\label{tab:main_results}
\end{table*}
Results in Table \ref{tab:main_results} underscore the effectiveness of PACED-RL in the mathematical reasoning domain. Crucially, PACED-RL consistently outperforms GRPO and FlowRL by a wide margin. On the Qwen2.5-Math-7B model, PACED-RL improves over the performance of GRPO and FlowRL by 29.1\% and 40.0\% respectively on the AIME 24 benchmark, providing strong evidence of the effectiveness of our method. PACED-RL also consistently matches or exceeds the performance of adaptive prompt selection baselines, despite not incorporating explicit auxiliary mechanisms for obtaining online accuracy estimates during training. This contrast is particularly notable given that  PACED-RL does not incur significant compute overhead from rollout generation, as with DS and LILO. The improvement is most pronounced on the hardest AIME benchmarks: for instance, on AIME 25 with Qwen2.5-Math-7B, PACED-RL attains 13.1 compared to 11.7 for LILO, corresponding to an 11.9\% relative gain.

\paragraph{Diversity.}
As a proxy for diversity, we assess the pass@$k$ performance of PACED-RL. Achieving high pass@$k$ requires the model to cover a broader space of plausible solution trajectories rather than concentrating probability mass on a single mode. Consequently, improvements in pass@$k$—particularly at larger $k$—are widely interpreted as evidence of increased solution diversity and exploration \citep{li2025jointly,chen2025pass,zhu2025the}. 

We compare the pass@$k$ performance of PACED-RL compared to GRPO, FlowRL, and the untrained base models Qwen2.5-Math-1.5B and 7B. We evaluate on the AIME24/25 benchmarks, and generate using temperature set to 0.6 and a top-$p$ value of 0.95, following \citet{yue2025does}. Fig.~\ref{fig:passk} shows that PACED-RL consistently outperforms baselines across values of $k$, with relative improvements over GRPO and FlowRL by up to 14.2\% and 9.1\% respectively. These results demonstrate the diversity-preserving nature and the superior exploration capability of PACED-RL.
\begin{figure}[ht]
  \centerline{\includegraphics[width=\columnwidth]{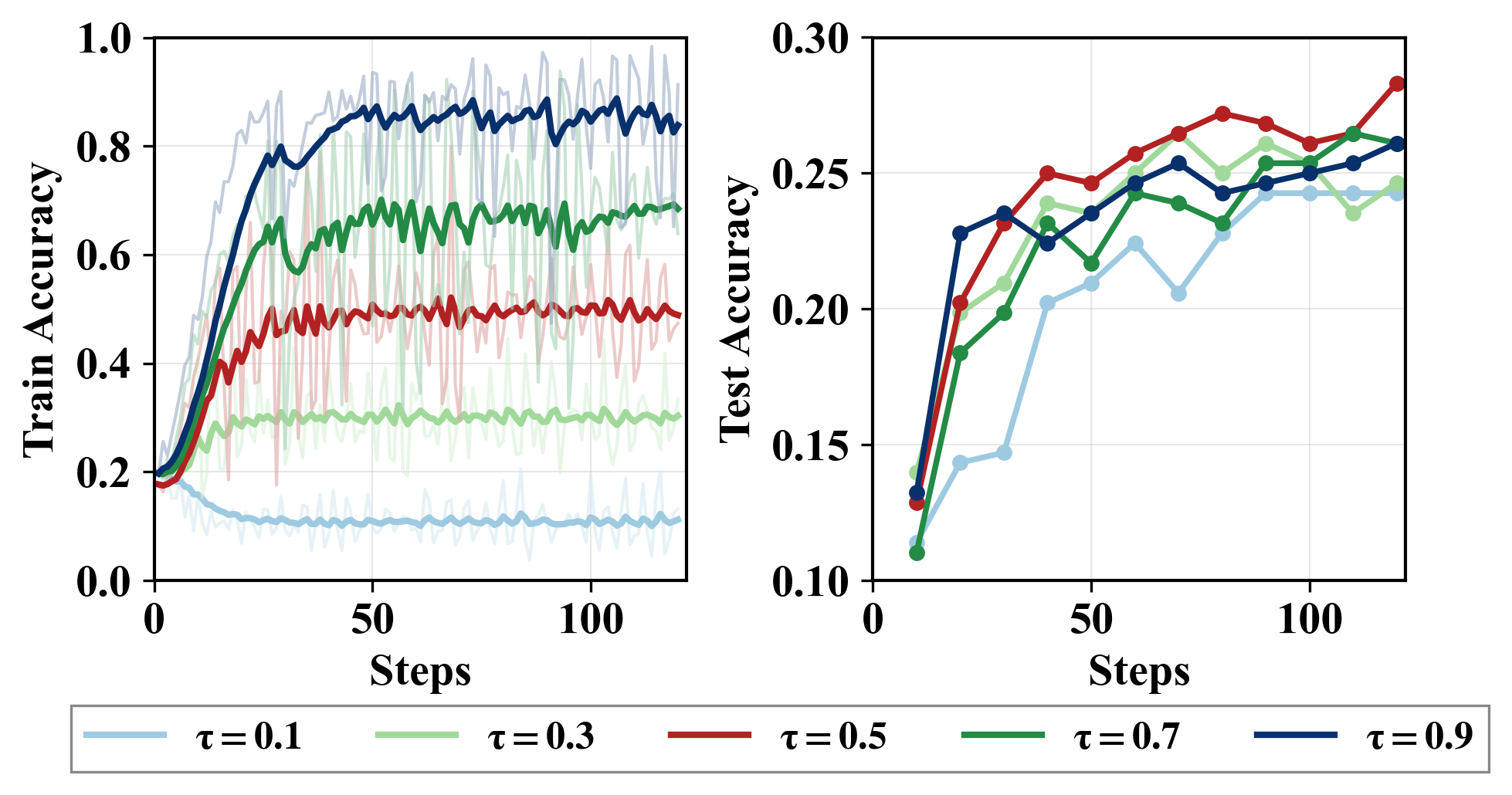}}
  \caption{Train and test accuracies of Qwen2.5-Math-1.5B trained on the DeepScaleR dataset and evaluated on the MinervaMath benchmark, for target accuracy values $\tau \in \{0.1,0.3,0.5,0.7,0.9\}$.}
  \label{fig:train_and_test}
\end{figure}

\subsection{Analyses}
\label{sec:analysis}
\paragraph{Controlling Train Accuracy with the Partition Function.}
We vary the target accuracy value $\tau$ to examine whether it enables control over the induced training accuracy and to assess its impact on the training dynamics. Fig.~\ref{fig:train_and_test} shows that our method selectively retains training questions whose estimated accuracies are close to $\tau$, thereby providing control over the training accuracy. Furthermore, our results reaffirm results from previous works \citep{foster2026lilo,gao2023scaling} that maintaining questions with accuracies closest to 0.5 is optimal for sample-efficiency, achieving the highest test accuracy throughout  training. Setting $\tau = 0.9$ yields the fastest \textit{initial} gains in test accuracy but quickly plateaus, ultimately resulting in sub-optimal performance. Moreover, the run with $\tau = 0.1$ produces the weakest results overall, with test accuracy consistently trailing all other settings throughout training. Taken together, these results indicate that by restricting training to an effective level of difficulty, adaptive prompt selection makes policy learning robust to dataset difficulty, enabling efficient and stable improvement on datasets of any difficulty via a sustained focus on the most informative question prompts.

\paragraph{On Discarding the KL Divergence Term.} We further empirically show that omitting the KL divergence term from the full expression in Eq.~\ref{eq:acc} to obtain the practical accuracy estimator in Eq.~\ref{eq:accuracy_final} does not significantly alter the sampling dynamics. Every 5 training steps, we randomly sample 512 prompts from $\mathcal{D}$ and compare the prompts selected by the two expressions.  Let $\mathcal{S}_{\mathrm{omit}}$ and $\mathcal{S}_{\mathrm{full}}$ denote the top-$m$ prompts whose estimated accuracies are closest to $0.5$ under Eq.~\ref{eq:accuracy_final} and Eq.~\ref{eq:acc}, respectively. We report the selection overlap $
\frac{|\mathcal{S}_{\mathrm{omit}} \cap \mathcal{S}_{\mathrm{full}}|}{k}
$, which ranges from 0 to 1, with 1 indicating identical selections. For Eq.~\ref{eq:acc}, the KL divergence term is obtained through Monte Carlo estimation using 8 rollouts, and as in the main experiments, we set $k=128$.

\begin{figure}[ht]
\hspace*{-0.018\columnwidth}{\includegraphics[width=\columnwidth]{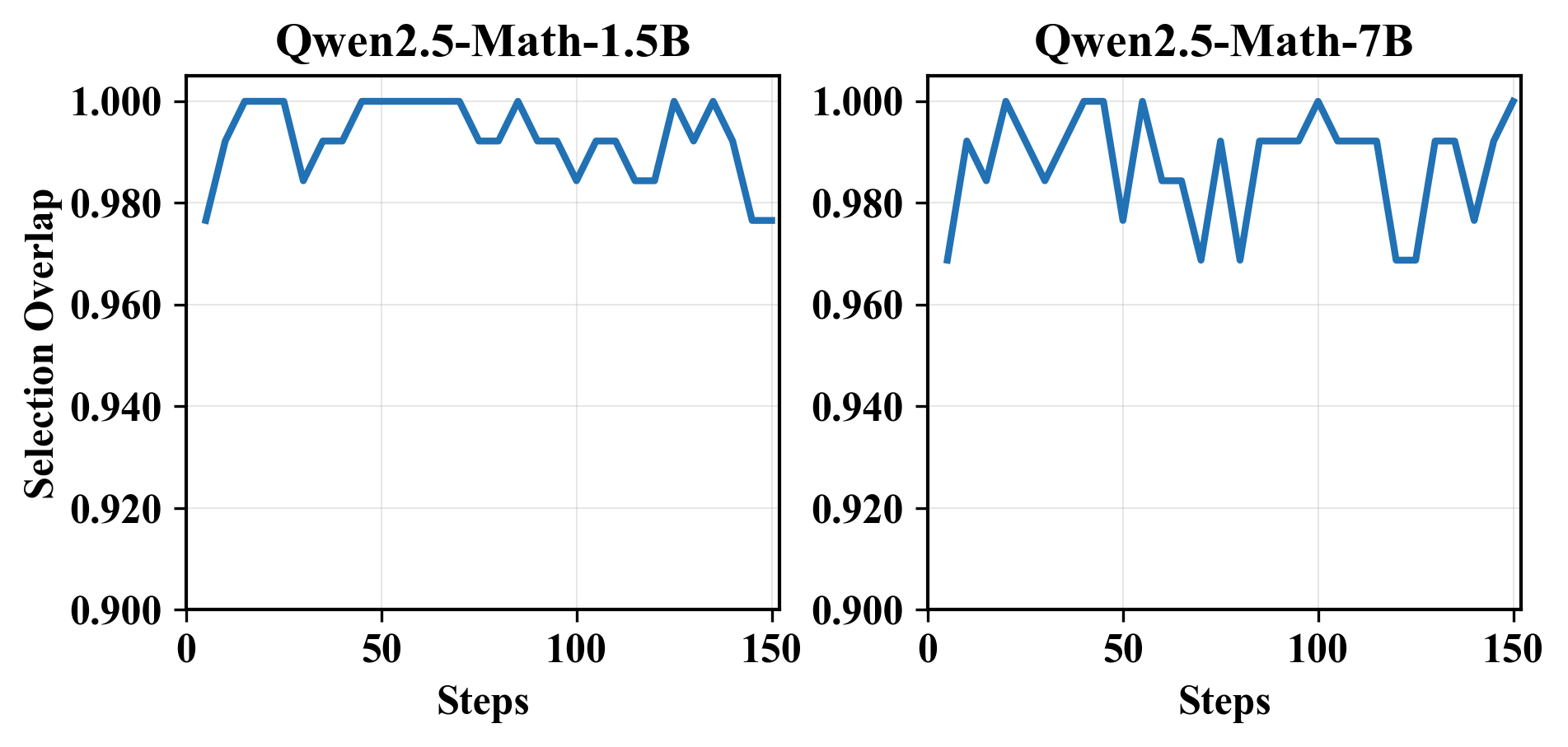}}
    \caption{
      Selection overlap between $\mathcal{S}_{\mathrm{omit}}$ and $\mathcal{S}_{\mathrm{full}}$ over the course of training. Results obtained using Qwen2.5-Math-1.5B and Qwen2.5-Math-7B models trained on the DeepScaleR dataset.
    }
    \label{fig:overlap}
\end{figure} 

Results in Figure.~\ref{fig:overlap} show that $\mathcal{S}_{\mathrm{omit}}$ and $\mathcal{S}_{\mathrm{full}}$ exhibit near-perfect overlap throughout training, indicating that omitting the KL divergence term has negligible effect on the sampling dynamics. Furthermore, estimation of the KL divergence term introduces substantial overhead, increasing training-step latency by 70\% for the 1.5B model and 87\% for the 7B model, despite being computed over only 512 prompts rather than the full dataset composed of 40k question prompts. These results illustrate that omitting the KL divergence term is a practical approximation for efficiently exploiting the partition function's accuracy signal, in the domain post-training LLMs for code and mathematial reasoning tasks.

\paragraph{Computational Overhead.}
Owing to the small size of the 3-layer MLP on top of pre-computed reference model
embeddings of question prompts $\mathbf{x}$ used to parameterize $Z_{\phi}$, the computation of $Z_{\phi}$ for all question prompts in the dataset can be efficiently performed using large-batch inference. To illustrate the negligible computational time overhead, we report in Table~\ref{tab:timing_breakdown} the average time required to compute $Z_{\phi}(\mathbf{x})$, and thus the accuracy estimates $\{\hat{p}_{\mathrm{old}}(\mathbf{x}_i)\}_{i=1}^{|\mathcal{D}|}$, alongside the average total runtime per training step. Across all models and training domains, the average time taken to obtain $Z_{\phi}(\mathbf{x})$ values is several orders of magnitude smaller than the average time taken per training step.
\begin{table}[ht]
\caption{Comparison between the average time taken for the computation of $Z_{\phi}(\mathbf{x})$ for all question prompts in the dataset and the average total time taken per step. DS-Distill-1.5B denotes the DeepSeek-R1-Distill-Qwen-1.5B model.}
\centering
\small
\setlength{\tabcolsep}{2pt}
\begin{tabular}{lcc}
\toprule
\textbf{Model} & \textbf{$Z_{\phi}(\mathbf{x})$ Computation (s)} & \textbf{Total (s)} \\
\midrule
Qwen2.5-Math-1.5B & 0.035 & 308 \\
Qwen2.5-Math-7B & 0.110 & 370 \\
DS-Distill-1.5B & 0.020 & 1086 \\
\bottomrule
\end{tabular}
\vspace{3pt}
\label{tab:timing_breakdown}
\end{table}

By exploiting these low-cost accuracy estimates obtainable with minimal additional latency, PACED-RL substantially improves sample efficiency, translating into notable performance gains.

\begin{figure}[ht]
    \centerline{\includegraphics[width=\columnwidth]{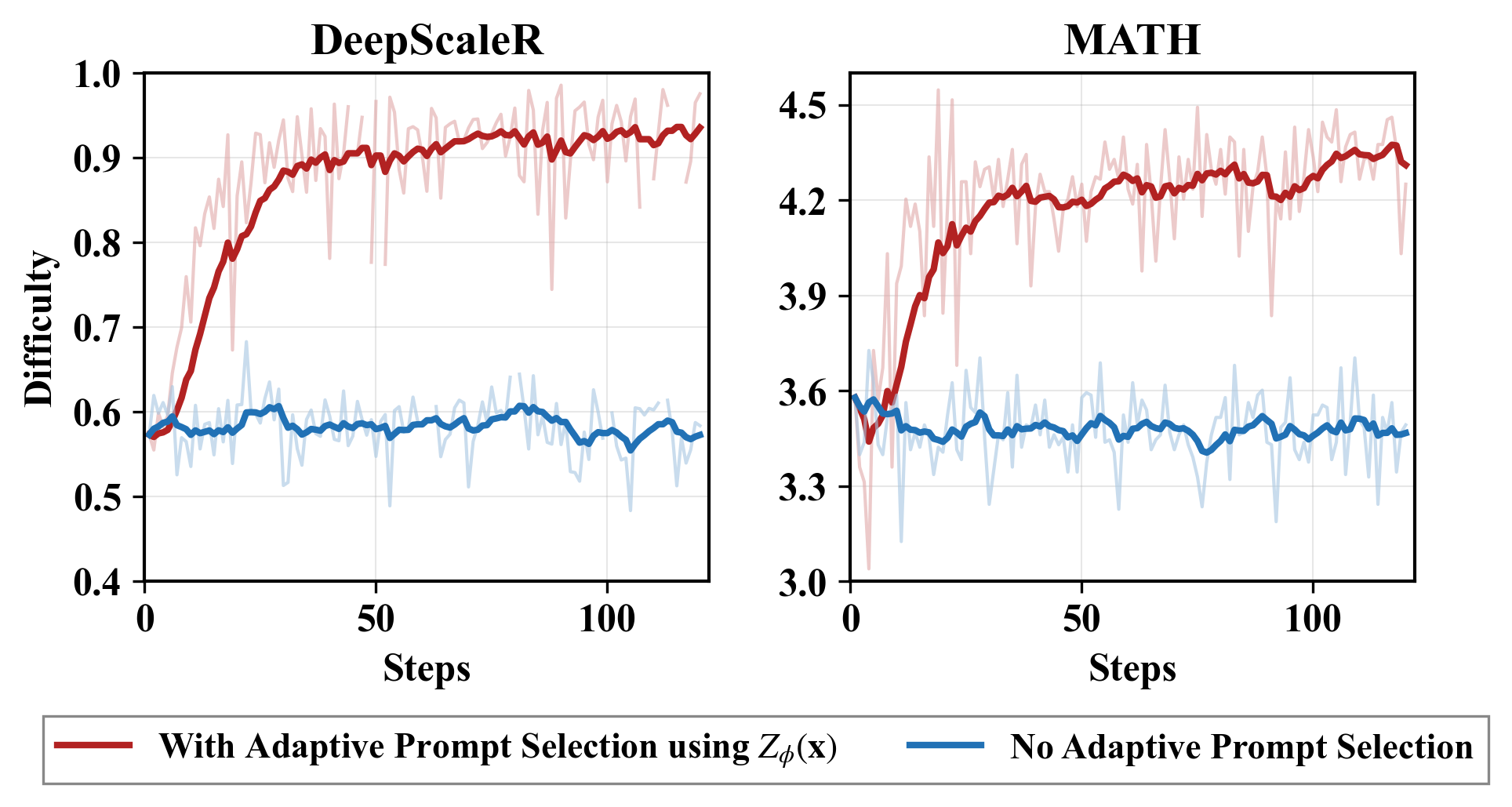}}
    \caption{
      Progression of the average difficulty of question prompts sampled in each training batch under the Qwen2.5-Math-7B model.
    }
    \label{fig:train_acc}
\end{figure}

\paragraph{Progression of Question Difficulty During Training.}
Figure~\ref{fig:train_acc} shows the evolution of the mean difficulty of sampled question prompts over the course of training. For the MATH dataset, we utilize the human-annotated difficulty labels, which range from level 1, indicating the simplest questions, to level 5, indicating the most complex questions. For DeepScaleR, which does not have human-annotated difficulty labels, we use the difficulty labels from \citet{shi2025efficientreinforcementfinetuningadaptive}, defined as 1 - pass@1 accuracy computed over all questions using the Qwen2.5-Math-7B model. Without adaptive prompt selection, the mean difficulty remains constant throughout training. However, with adaptive prompt selection guided by partition function–guided accuracy estimates, the mean difficulties progressively increase, shifting toward harder questions as the model improves. At latter stages of training, PACED-RL selectively trains on the hardest subsets of the dataset. This trend is expected, as a question regarded as having intermediate difficulty in earlier parts of training is likely to become easier for the policy as training progresses.
\begin{table}[ht]
\caption{Relative coverage and total number of distinct prompts observed during 150 training steps for the Qwen2.5-Math-1.5B and 7B models trained on the DeepScaleR dataset.}
\centering
\small
\setlength{\tabcolsep}{2pt}
\begin{tabular}{lcc}
\toprule
\textbf{Metric} & \textbf{Qwen2.5-Math-1.5B} & \textbf{Qwen2.5-Math-7B} \\
\midrule
Relative Coverage & 55.7\% & 62.3\% \\
Distinct Prompts & 10695 & 11962 \\
\bottomrule
\end{tabular}
\vspace{3pt}
\label{tab:coverage_distinct_prompts}
\end{table}

Such progression of question difficulty naturally prevents PACED-RL from collapsing to a small subset of the dataset during training, despite it greedily selecting the most informative prompts at each training step. We additionally report \textit{relative coverage}, defined as the ratio of unique prompts observed by PACED-RL relative to the no-adaptive-prompt-selection variant, together with the total number of distinct prompts encountered during training. As shown in Table.~\ref{tab:coverage_distinct_prompts}, PACED-RL trains on a sizeable subset of the prompts seen by the no-adaptive-prompt-selection variant, filtering out less informative prompts to avoid unnecessary trajectory generation and optimization. Moreover, the number of distinct prompts remains large ($>10$k), indicating that PACED-RL does not collapse to a narrow subset of the dataset.
\paragraph{Ablation on the Sampling Mechanism.} PACED-RL greedily selects the top-$m$ prompts whose estimated accuracies are the closest to $0.5$ to maximize sample efficiency. To assess the sensitivity of the training dynamics of PACED-RL to this deterministic prompt selection rule, we additionally evaluate a soft-sampling variant. Specifically, we replace the greedy top-$m$ selection with sampling from a softmax distribution whose logits are defined as $\hat{p}_{\mathrm{old}}(\mathbf{x})(1-\hat{p}_{\mathrm{old}}(\mathbf{x}))$, with temperature $T$. Since the logits are maximized at $\hat{p}_{\mathrm{old}}(\mathbf{x})=0.5$, this configuration naturally assigns higher sampling probability to prompts with estimated accuracies closer to $0.5$,  while allowing prompts in a broader neighborhood around $0.5$ to be sampled.
\begin{table}[ht]
\caption{MATH-500 Avg@8 performance using different prompt sampling strategies. Results obtained with the Qwen2.5-Math-1.5B model trained on the DeepScaleR dataset.}
\centering
\small
\setlength{\tabcolsep}{3pt}
\begin{tabular}{lcccc}
\toprule
\textbf{Temperature} & \textbf{\boldmath$T=1.0$} & \textbf{\boldmath$T=0.7$} & \textbf{\boldmath$T=0.4$} & \textbf{Greedy} \\
\midrule
MATH-500 Avg@8 & 72.2 & 71.4 & 71.8 & 73.2 \\
\bottomrule
\end{tabular}
\vspace{3pt}
\label{tab:selection_strategy_temperature}
\end{table}

Table.~\ref{tab:selection_strategy_temperature} reports Avg@8 performance on MATH500 for $T \in \{0.4,0.7,1.0\}$. We observe that soft sampling consistently underperforms the default greedy top-$m$ selection. One possible explanation is that soft sampling assigns non-negligible probability to less informative prompts, leading to such prompts being inserted into the training batch. The quality of the training signal may thus be reduced, and could lead to weaker downstream performances.
\begin{table*}[t]
\caption{Ablation on the accuracy-estimation-error-prioritized replay for Qwen2.5-Math-1.5B trained on the DeepScaleR training set. We report the pass@1 performance across mathematical reasoning benchmarks, with Avg@32 for the AIME24 and AIME25 benchmarks. \textbf{Avg.} denotes the average of the performance on each of the benchmarks.}
\centering
\footnotesize
\begin{adjustbox}{max width=\textwidth,center}
\begin{tabular}{llcccccc}
\toprule
\textbf{Model}
& \textbf{Method} 
& \textbf{MATH-500}
& \textbf{Olympiad}
& \textbf{Minerva} 
& \makecell{\textbf{AIME 24}\\\textbf{Avg@32}} 
& \makecell{\textbf{AIME 25}\\\textbf{Avg@32}} 
& \textbf{Avg.}
\\
\midrule

\multirow[c]{1}{*}{\centering Qwen2.5-Math-1.5B}
& \cellcolor{gray!15}PACED-RL
& \cellcolor{gray!15}73.2
& \cellcolor{gray!15}37.9
& \cellcolor{gray!15}29.0
& \cellcolor{gray!15}13.3
& \cellcolor{gray!15}7.3
& \cellcolor{gray!15}32.1
\\
& \hspace{1em}w/o Replay
& 73.0 & 36.3 & 26.1 & 9.8 & 6.1 & 30.2
\\

\bottomrule
\end{tabular}
\end{adjustbox}
\vspace{3pt}
\label{tab:replay}
\end{table*}
\paragraph{Ablation on the Replay Mechanism.}
To isolate the contribution of our accuracy estimate error–prioritized replay mechanism, we conduct an ablation study on the replay strategy in Table \ref{tab:replay}. Incorporating replay yields consistent performance improvements across all mathematical reasoning benchmarks, indicating that selectively revisiting question–output pairs whose predicted accuracy disagrees largely with observed accuracy provides an informative training signal for improving sample efficiency.

\begin{figure}[ht]
\centerline{\includegraphics[width=0.8\columnwidth]{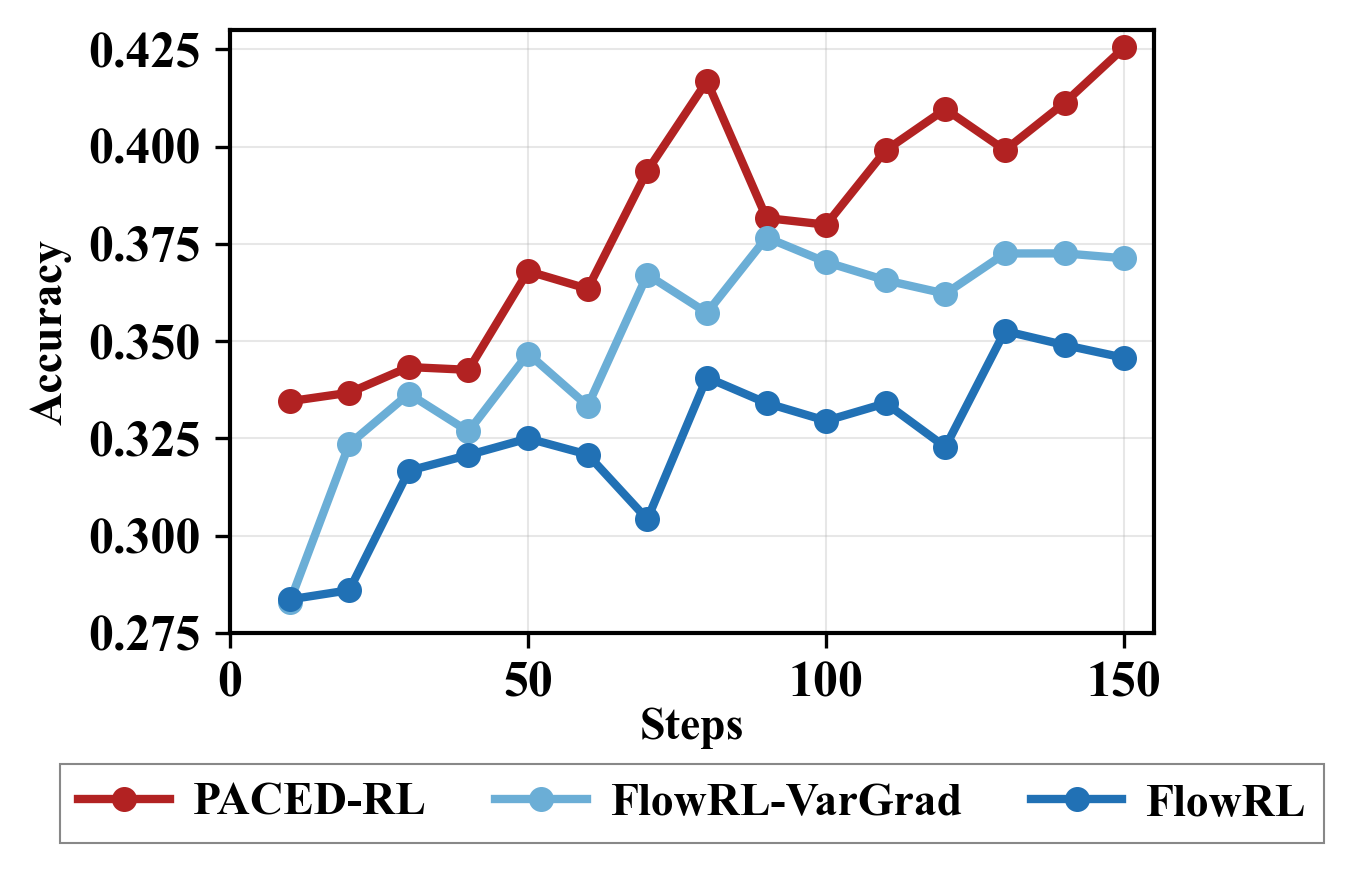}}
    \caption{
      Progression of average test accuracy of FlowRL, FlowRL-VarGrad, and PACED-RL across the 5 mathematical reasoning benchmarks. Results obtained using Qwen2.5-Math-1.5B trained on the DeepScaleR dataset.
    }
    \label{fig:vargrad}
\end{figure} 
\paragraph{Comparison to TB-VarGrad.}
To stabilize training by circumventing the need to train the partition function, several prior works \citep{zhang2023robust,venkatraman2024amortizing,bartoldson2025trajectory} adopt the VarGrad variant of the Trajectory Balance objective, which replaces the learned partition function $Z_{\phi}(\mathbf{x})$ with a batch-level estimate. Specifically, for a prompt $\mathbf{x}_i$ and its $N$ corresponding rollouts $\{\mathbf{y}_{i,j}\}_{j=1}^{N}$, the batch estimate of $Z_{\phi}(\mathbf{x}_i)$ can be obtained as:
\begin{equation}
\label{eq:vargrad}
\log \hat{Z}\!\left(\mathbf{x}_{i}\right)
=
\frac{1}{N}
\sum_{j=1}^{N}
\left(
R(\mathbf{x}_i,\mathbf{y}_{i,j})
-
\log \pi_{\theta}\!\left(\mathbf{y}_{i,j} \mid \mathbf{x}_{i}\right)
\right).
\end{equation}

Crucially, as the estimate in Eq. \ref{eq:vargrad} can only be made \textit{after} the rollout generation phase, it cannot be used as an online accuracy estimator to guide adaptive prompt selection.

Using the Qwen2.5-Math-1.5B model, we compare PACED-RL against FlowRL-VarGrad, a variant of FlowRL that uses the batch-estimate $Z_{\phi}(\mathbf{x})$ for training instead of learning the partition function. Consistent with the observations of \citet{zhang2023robust}, Fig.~\ref{fig:vargrad} shows that FlowRL-VarGrad converges faster and achieves higher test performance than the original FlowRL approach. Nevertheless, PACED-RL consistently outperforms FlowRL-VarGrad throughout training, demonstrating that explicitly learning the partition function—and leveraging the accuracy information it encodes for increased sample efficiency—is more effective than relying on a batch-estimate approximation.

\section{Conclusion \& Future Works}
\paragraph{Conclusion.}
We introduce PACED-RL, a distribution-matching post-training framework that improves LLM reasoning performances by explicitly leveraging the learned partition function as an online accuracy signal. By exploiting the accuracy signals obtained without incurring additional compute cost, PACED-RL adaptively concentrates training resources on the most informative training samples, leading to a more effective and sample-efficient RLVR post-training for LLM reasoning. Extensive experiments on mathematical reasoning and code generation tasks across three models demonstrate the consistent effectiveness of our proposed approach. Overall, by reinterpreting the partition function as an online accuracy signal, PACED-RL offers a principled pathway towards more sample-efficient distribution-matching methods for training LLMs.

\paragraph{Future Works}
The interpretation of the partition function as an online accuracy estimator opens up new opportunities for inference time strategies. Beyond its role in training, future work could investigate techniques such as adaptive self-consistency, where additional samples are selectively allocated to question prompts predicted to be difficult according to the learned partition function, enabling more compute-efficient and accuracy-aware reasoning.

Additionally, we leave the study of applying GFlowNet objective variants—such as Detailed Balance \citep{bengio2023gflownet} and Sub-Trajectory Balance \citep{madan2023learning}—to PACED-RL in multi-step reasoning settings for future work. This extension would allow us to transition from leveraging the partition function as a question-level difficulty estimator to utilizing intermediate state \textit{flows} as step-level difficulty signals, thereby enabling more granular credit assignment and difficulty-aware guidance in multi-step reasoning tasks.

Lastly, a promising direction for future works is to extend PACED-RL to asynchronous GFlowNet reinforcement learning frameworks \citep{bartoldson2025trajectory} to substantially improve scalability and training throughput.

\section*{Acknowledgments}
This work was partly supported by Institute of Information \& communications Technology Planning \& Evaluation (IITP) grant funded by the Korea government (MSIT) [No.RS-2022-II220184, Development and Study of AI Technologies to Inexpensively Conform to Evolving Policy on Ethics \& No.RS-2021-II212068, Artificial Intelligence Innovation Hub (Artificial Intelligence Institute, Seoul National University)]. K. Jung is with ASRI, Seoul National University, Korea.
\section*{Impact Statement}
This paper presents work whose goal is to advance the field of Machine
Learning. We confirm that all datasets included in our study are sourced from established, publicly available repositories and standard benchmarks. There are many potential societal consequences of our work, none
which we feel must be specifically highlighted here.


\bibliography{example_paper}

@inproceedings{
zhu2026flowrl,
title={Flow{RL}: Matching Reward Distributions for {LLM} Reasoning},
author={Xuekai Zhu and Daixuan Cheng and Dinghuai Zhang and Hengli Li and Kaiyan Zhang and Che Jiang and Youbang Sun and Ermo Hua and Yuxin Zuo and Xingtai Lv and Qizheng Zhang and Lin Chen and Fanghao Shao and Bo Xue and Yunchong Song and Zhenjie Yang and Ganqu Cui and Ning Ding and Jianfeng Gao and Xiaodong Liu and Bowen Zhou and Hongyuan Mei and Zhouhan Lin},
booktitle={The Fourteenth International Conference on Learning Representations},
year={2026},
url={https://openreview.net/forum?id=lObnTKbm9U}
}

@inproceedings{
bartoldson2025trajectory,
title={Trajectory Balance with Asynchrony: Decoupling Exploration and Learning for Fast, Scalable {LLM} Post-Training},
author={Brian R. Bartoldson and Siddarth Venkatraman and James Diffenderfer and Moksh Jain and Tal Ben-Nun and Seanie Lee and Minsu Kim and Johan Obando-Ceron and Yoshua Bengio and Bhavya Kailkhura},
booktitle={The Thirty-ninth Annual Conference on Neural Information Processing Systems},
year={2025},
url={https://openreview.net/forum?id=VwPt1WDQNB}
}

@article{bengio2023gflownet,
  title={Gflownet foundations},
  author={Bengio, Yoshua and Lahlou, Salem and Deleu, Tristan and Hu, Edward J and Tiwari, Mo and Bengio, Emmanuel},
  journal={Journal of Machine Learning Research},
  volume={24},
  number={210},
  pages={1--55},
  year={2023}
}

@inproceedings{
lee2025learning,
title={Learning Diverse Attacks on Large Language Models for Robust Red-Teaming and Safety Tuning},
author={Seanie Lee and Minsu Kim and Lynn Cherif and David Dobre and Juho Lee and Sung Ju Hwang and Kenji Kawaguchi and Gauthier Gidel and Yoshua Bengio and Nikolay Malkin and Moksh Jain},
booktitle={The Thirteenth International Conference on Learning Representations},
year={2025},
url={https://openreview.net/forum?id=1mXufFuv95}
}

@article{venkatraman2024amortizing,
  title={Amortizing intractable inference in diffusion models for vision, language, and control},
  author={Venkatraman, Siddarth and Jain, Moksh and Scimeca, Luca and Kim, Minsu and Sendera, Marcin and Hasan, Mohsin and Rowe, Luke and Mittal, Sarthak and Lemos, Pablo and Bengio, Emmanuel and others},
  journal={Advances in neural information processing systems},
  volume={37},
  pages={76080--76114},
  year={2024}
}

@inproceedings{kwon2024gdpo,
  title={GDPO: Learning to Directly Align Language Models with Diversity Using GFlowNets},
  author={Kwon, Oh Joon and Matsunaga, Daiki and Kim, Kee-Eung},
  booktitle={Proceedings of the 2024 Conference on Empirical Methods in Natural Language Processing},
  pages={17120--17139},
  year={2024}
}

@inproceedings{zhang2023robust,
  title={Robust Scheduling with GFlowNets},
  author={Zhang, David W and Rainone, Corrado and Peschl, Markus and Bondesan, Roberto},
  booktitle={ICLR},
  year={2023}
}

@inproceedings{
yu2025flow,
title={Flow of Reasoning: Training {LLM}s for Divergent Reasoning with Minimal Examples},
author={Fangxu Yu and Lai Jiang and Haoqiang Kang and Shibo Hao and Lianhui Qin},
booktitle={Forty-second International Conference on Machine Learning},
year={2025},
url={https://openreview.net/forum?id=qyMxunrR2j}
}

@article{shao2024deepseekmath,
  title={Deepseekmath: Pushing the limits of mathematical reasoning in open language models},
  author={Shao, Zhihong and Wang, Peiyi and Zhu, Qihao and Xu, Runxin and Song, Junxiao and Bi, Xiao and Zhang, Haowei and Zhang, Mingchuan and Li, YK and others},
  journal={arXiv preprint arXiv:2402.03300},
  year={2024}
}

@article{malkin2022trajectory,
  title={Trajectory balance: Improved credit assignment in gflownets},
  author={Malkin, Nikolay and Jain, Moksh and Bengio, Emmanuel and Sun, Chen and Bengio, Yoshua},
  journal={Advances in Neural Information Processing Systems},
  volume={35},
  pages={5955--5967},
  year={2022}
}

@article{yu2025dapo,
  title={Dapo: An open-source llm reinforcement learning system at scale},
  author={Yu, Qiying and Zhang, Zheng and Zhu, Ruofei and Yuan, Yufeng and Zuo, Xiaochen and Yue, Yu and Dai, Weinan and Fan, Tiantian and Liu, Gaohong and Liu, Lingjun and others},
  journal={arXiv preprint arXiv:2503.14476},
  year={2025}
}

@article{foster2026lilo,
  title={LILO: Learning to Reason at the Frontier of Learnability},
  author={Foster, Thomas and Sims, Anya and Forkel, Johannes and Foerster, Jakob},
  journal={Advances in Neural Information Processing Systems},
  volume={38},
  pages={48941--48974},
  year={2026}
}

@inproceedings{zhang2025learning,
  title={Learning like humans: Advancing llm reasoning capabilities via adaptive difficulty curriculum learning and expert-guided self-reformulation},
  author={Zhang, Enci and Yan, Xingang and Lin, Wei and Zhang, Tianxiang and Qianchun, Lu},
  booktitle={Proceedings of the 2025 Conference on Empirical Methods in Natural Language Processing},
  pages={6630--6644},
  year={2025}
}

@inproceedings{
zheng2025act,
title={Act Only When It Pays: Efficient Reinforcement Learning for {LLM} Reasoning via Selective Rollouts},
author={Haizhong Zheng and Yang Zhou and Brian R. Bartoldson and Bhavya Kailkhura and Fan Lai and Jiawei Zhao and Beidi Chen},
booktitle={The Thirty-ninth Annual Conference on Neural Information Processing Systems},
year={2025},
url={https://openreview.net/forum?id=x5lITYXmW2}
}

@inproceedings{qu2026can,
  title={Can prompt difficulty be online predicted for accelerating rl finetuning of reasoning models?},
  author={Qu, Yun and Wang, Qi and Mao, Yixiu and Hu, Vincent Tao and Ommer, Bj{\"o}rn and Ji, Xiangyang},
  booktitle={Proceedings of the 32nd ACM SIGKDD Conference on Knowledge Discovery and Data Mining V. 1},
  pages={1240--1250},
  year={2026}
}

@inproceedings{bae-etal-2026-online,
    title = "Online Difficulty Filtering for Reasoning Oriented Reinforcement Learning",
    author = "Bae, Sanghwan  and
      Hong, Jiwoo  and
      Lee, Min Young  and
      Kim, Hanbyul  and
      Nam, Jeongyeon  and
      Kwak, Donghyun",
    editor = "Demberg, Vera  and
      Inui, Kentaro  and
      Marquez, Llu{\'i}s",
    booktitle = "Proceedings of the 19th Conference of the {E}uropean Chapter of the {A}ssociation for {C}omputational {L}inguistics (Volume 1: Long Papers)",
    month = mar,
    year = "2026",
    address = "Rabat, Morocco",
    publisher = "Association for Computational Linguistics",
    url = "https://aclanthology.org/2026.eacl-long.30/",
    doi = "10.18653/v1/2026.eacl-long.30",
    pages = "700--719",
    ISBN = "979-8-89176-380-7"
}

@article{martens2020new,
  title={New insights and perspectives on the natural gradient method},
  author={Martens, James},
  journal={Journal of Machine Learning Research},
  volume={21},
  number={146},
  pages={1--76},
  year={2020}
}

@inproceedings{gao2023scaling,
  title={Scaling laws for reward model overoptimization},
  author={Gao, Leo and Schulman, John and Hilton, Jacob},
  booktitle={International Conference on Machine Learning},
  pages={10835--10866},
  year={2023},
  organization={PMLR}
}

@inproceedings{
padmakumar2024does,
title={Does Writing with Language Models Reduce Content Diversity?},
author={Vishakh Padmakumar and He He},
booktitle={The Twelfth International Conference on Learning Representations},
year={2024},
url={https://openreview.net/forum?id=Feiz5HtCD0}
}

@inproceedings{
shypula2025evaluating,
title={Evaluating the Diversity and Quality of {LLM} Generated Content},
author={Alexander Shypula and Shuo Li and Botong Zhang and Vishakh Padmakumar and Kayo Yin and Osbert Bastani},
booktitle={Second Conference on Language Modeling},
year={2025},
url={https://openreview.net/forum?id=O7bF6nlSOD}
}

@article{li2025jointly,
  title={Jointly reinforcing diversity and quality in language model generations},
  author={Li, Tianjian and Zhang, Yiming and Yu, Ping and Saha, Swarnadeep and Khashabi, Daniel and Weston, Jason and Lanchantin, Jack and Wang, Tianlu},
  journal={arXiv preprint arXiv:2509.02534},
  year={2025}
}

@inproceedings{
zhu2025the,
title={The Surprising Effectiveness of Negative Reinforcement in {LLM} Reasoning},
author={Xinyu Zhu and Mengzhou Xia and Zhepei Wei and Wei-Lin Chen and Danqi Chen and Yu Meng},
booktitle={The Thirty-ninth Annual Conference on Neural Information Processing Systems},
year={2025},
url={https://openreview.net/forum?id=ftVlLG9cks}
}

@article{schulman2017proximal,
  title={Proximal policy optimization algorithms},
  author={Schulman, John and Wolski, Filip and Dhariwal, Prafulla and Radford, Alec and Klimov, Oleg},
  journal={arXiv preprint arXiv:1707.06347},
  year={2017}
}

@article{schaul2015prioritized,
  title={Prioritized experience replay},
  author={Schaul, Tom and Quan, John and Antonoglou, Ioannis and Silver, David},
  journal={arXiv preprint arXiv:1511.05952},
  year={2015}
}

@inproceedings{shen2023towards,
  title={Towards understanding and improving gflownet training},
  author={Shen, Max W and Bengio, Emmanuel and Hajiramezanali, Ehsan and Loukas, Andreas and Cho, Kyunghyun and Biancalani, Tommaso},
  booktitle={International conference on machine learning},
  pages={30956--30975},
  year={2023},
  organization={PMLR}
}

@inproceedings{
kim2024local,
title={Local Search {GF}lowNets},
author={Minsu Kim and Taeyoung Yun and Emmanuel Bengio and Dinghuai Zhang and Yoshua Bengio and Sungsoo Ahn and Jinkyoo Park},
booktitle={The Twelfth International Conference on Learning Representations},
year={2024},
url={https://openreview.net/forum?id=6cFcw1Rxww}
}

@inproceedings{achiam2017constrained,
  title={Constrained policy optimization},
  author={Achiam, Joshua and Held, David and Tamar, Aviv and Abbeel, Pieter},
  booktitle={International conference on machine learning},
  pages={22--31},
  year={2017},
  organization={PMLR}
}

@inproceedings{
zeng2026cures,
title={Cur{ES}: From Gradient Analysis to Efficient Curriculum Learning for Reasoning {LLM}s},
author={Yongcheng Zeng and Zexu Sun and Bokai Ji and Erxue Min and Hengyi Cai and Shuaiqiang Wang and Dawei Yin and Haifeng Zhang and Xu Chen and Jun Wang},
booktitle={The Fourteenth International Conference on Learning Representations},
year={2026},
url={https://openreview.net/forum?id=QXrZ0Y3yGJ}
}

@article{yang2024qwen2,
  title={Qwen2. 5-math technical report: Toward mathematical expert model via self-improvement},
  author={Yang, An and Zhang, Beichen and Hui, Binyuan and Gao, Bofei and Yu, Bowen and Li, Chengpeng and Liu, Dayiheng and Tu, Jianhong and Zhou, Jingren and Lin, Junyang and others},
  journal={arXiv preprint arXiv:2409.12122},
  year={2024}
}

@inproceedings{
hendrycks2021measuring,
title={Measuring Mathematical Problem Solving With the {MATH} Dataset},
author={Dan Hendrycks and Collin Burns and Saurav Kadavath and Akul Arora and Steven Basart and Eric Tang and Dawn Song and Jacob Steinhardt},
booktitle={Thirty-fifth Conference on Neural Information Processing Systems Datasets and Benchmarks Track (Round 2)},
year={2021},
url={https://openreview.net/forum?id=7Bywt2mQsCe}
}

@article{luo2025deepscaler,
  title={Deepscaler: Surpassing o1-preview with a 1.5 b model by scaling rl},
  author={Luo, Michael and Tan, Sijun and Wong, Justin and Shi, Xiaoxiang and Tang, William Y and Roongta, Manan and Cai, Colin and Luo, Jeffrey and Zhang, Tianjun and Li, Li Erran and others},
  journal={Notion Blog},
  year={2025}
}

@inproceedings{sheng2025hybridflow,
  title={Hybridflow: A flexible and efficient rlhf framework},
  author={Sheng, Guangming and Zhang, Chi and Ye, Zilingfeng and Wu, Xibin and Zhang, Wang and Zhang, Ru and Peng, Yanghua and Lin, Haibin and Wu, Chuan},
  booktitle={Proceedings of the Twentieth European Conference on Computer Systems},
  pages={1279--1297},
  year={2025}
}

@article{lewkowycz2022solving,
  title={Solving quantitative reasoning problems with language models},
  author={Lewkowycz, Aitor and Andreassen, Anders and Dohan, David and Dyer, Ethan and Michalewski, Henryk and Ramasesh, Vinay and Slone, Ambrose and Anil, Cem and Schlag, Imanol and Gutman-Solo, Theo and others},
  journal={Advances in neural information processing systems},
  volume={35},
  pages={3843--3857},
  year={2022}
}

@inproceedings{he2024olympiadbench,
  title={Olympiadbench: A challenging benchmark for promoting agi with olympiad-level bilingual multimodal scientific problems},
  author={He, Chaoqun and Luo, Renjie and Bai, Yuzhuo and Hu, Shengding and Thai, Zhen and Shen, Junhao and Hu, Jinyi and Han, Xu and Huang, Yujie and Zhang, Yuxiang and others},
  booktitle={Proceedings of the 62nd Annual Meeting of the Association for Computational Linguistics (Volume 1: Long Papers)},
  pages={3828--3850},
  year={2024}
}

@misc{shi2025efficientreinforcementfinetuningadaptive,
  title={Efficient Reinforcement Finetuning via Adaptive Curriculum Learning}, 
  author={Taiwei Shi and Yiyang Wu and Linxin Song and Tianyi Zhou and Jieyu Zhao},
  year={2025},
  eprint={2504.05520},
  archivePrefix={arXiv},
  primaryClass={cs.LG},
  url={https://arxiv.org/abs/2504.05520}, 
}

@inproceedings{
yue2025does,
title={Does Reinforcement Learning Really Incentivize Reasoning Capacity in {LLM}s Beyond the Base Model?},
author={Yang Yue and Zhiqi Chen and Rui Lu and Andrew Zhao and Zhaokai Wang and Yang Yue and Shiji Song and Gao Huang},
booktitle={The Thirty-ninth Annual Conference on Neural Information Processing Systems},
year={2025},
url={https://openreview.net/forum?id=4OsgYD7em5}
}

@article{luo2025deepcoder,
  title={Deepcoder: A fully open-source 14b coder at o3-mini level},
  author={Luo, Michael and Tan, Sijun and Huang, Roy and Patel, Ameen and Ariyak, Alpay and Wu, Qingyang and Shi, Xiaoxiang and Xin, Rachel and Cai, Colin and Weber, Maurice and others},
  journal={Notion Blog},
  year={2025}
}

@inproceedings{
jain2025livecodebench,
title={LiveCodeBench: Holistic and Contamination Free Evaluation of Large Language Models for Code},
author={Naman Jain and King Han and Alex Gu and Wen-Ding Li and Fanjia Yan and Tianjun Zhang and Sida Wang and Armando Solar-Lezama and Koushik Sen and Ion Stoica},
booktitle={The Thirteenth International Conference on Learning Representations},
year={2025},
url={https://openreview.net/forum?id=chfJJYC3iL}
}

@article{chen2021evaluating,
  title={Evaluating large language models trained on code},
  author={Chen, Mark},
  journal={arXiv preprint arXiv:2107.03374},
  year={2021}
}

@article{guo2025deepseek,
  title={Deepseek-r1: Incentivizing reasoning capability in llms via reinforcement learning},
  author={Guo, Daya and Yang, Dejian and Zhang, Haowei and Song, Junxiao and Zhang, Ruoyu and Xu, Runxin and Zhu, Qihao and Ma, Shirong and Wang, Peiyi and Bi, Xiao and others},
  journal={arXiv preprint arXiv:2501.12948},
  year={2025}
}

@inproceedings{
huang2025selfimprovement,
title={Self-Improvement in Language Models: The Sharpening Mechanism},
author={Audrey Huang and Adam Block and Dylan J Foster and Dhruv Rohatgi and Cyril Zhang and Max Simchowitz and Jordan T. Ash and Akshay Krishnamurthy},
booktitle={The Thirteenth International Conference on Learning Representations},
year={2025},
url={https://openreview.net/forum?id=WJaUkwci9o}
}

@inproceedings{
li2025preserving,
title={Preserving Diversity in Supervised Fine-Tuning of Large Language Models},
author={Ziniu Li and Congliang Chen and Tian Xu and Zeyu Qin and Jiancong Xiao and Zhi-Quan Luo and Ruoyu Sun},
booktitle={The Thirteenth International Conference on Learning Representations},
year={2025},
url={https://openreview.net/forum?id=NQEe7B7bSw}
}

@inproceedings{
horgan2018distributed,
title={Distributed Prioritized Experience Replay},
author={Dan Horgan and John Quan and David Budden and Gabriel Barth-Maron and Matteo Hessel and Hado van Hasselt and David Silver},
booktitle={International Conference on Learning Representations},
year={2018},
url={https://openreview.net/forum?id=H1Dy---0Z},
}

@inproceedings{hessel2018rainbow,
  title={Rainbow: Combining improvements in deep reinforcement learning},
  author={Hessel, Matteo and Modayil, Joseph and Van Hasselt, Hado and Schaul, Tom and Ostrovski, Georg and Dabney, Will and Horgan, Dan and Piot, Bilal and Azar, Mohammad and Silver, David},
  booktitle={Proceedings of the AAAI conference on artificial intelligence},
  volume={32},
  number={1},
  year={2018}
}

@article{sujit2023prioritizing,
  title={Prioritizing samples in reinforcement learning with reducible loss},
  author={Sujit, Shivakanth and Nath, Somjit and Braga, Pedro and Ebrahimi Kahou, Samira},
  journal={Advances in Neural Information Processing Systems},
  volume={36},
  pages={23237--23258},
  year={2023}
}

@inproceedings{fedus2020revisiting,
  title={Revisiting fundamentals of experience replay},
  author={Fedus, William and Ramachandran, Prajit and Agarwal, Rishabh and Bengio, Yoshua and Larochelle, Hugo and Rowland, Mark and Dabney, Will},
  booktitle={International conference on machine learning},
  pages={3061--3071},
  year={2020},
  organization={PMLR}
}

@article{wang2025eframe,
  title={EFRame: Deeper Reasoning via Exploration-Filtering-Replay Reinforcement Learning Framework},
  author={Wang, Chen and Wei, Lai and Zhang, Yanzhi and Shao, Chenyang and Dan, Zedong and Huang, Weiran and Wang, Yue and Zhang, Yuzhi},
  journal={arXiv preprint arXiv:2506.22200},
  year={2025}
}

@article{li2025repo,
  title={RePO: Replay-Enhanced Policy Optimization},
  author={Li, Siheng and Zhou, Zhanhui and Lam, Wai and Yang, Chao and Lu, Chaochao},
  journal={arXiv preprint arXiv:2506.09340},
  year={2025}
}

@article{chen2025pass,
  title={Pass@ k training for adaptively balancing exploration and exploitation of large reasoning models},
  author={Chen, Zhipeng and Qin, Xiaobo and Wu, Youbin and Ling, Yue and Ye, Qinghao and Zhao, Wayne Xin and Shi, Guang},
  journal={arXiv preprint arXiv:2508.10751},
  year={2025}
}

@article{comanici2025gemini,
  title={Gemini 2.5: Pushing the frontier with advanced reasoning, multimodality, long context, and next generation agentic capabilities},
  author={Comanici, Gheorghe and Bieber, Eric and Schaekermann, Mike and Pasupat, Ice and Sachdeva, Noveen and Dhillon, Inderjit and Blistein, Marcel and Ram, Ori and Zhang, Dan and Rosen, Evan and others},
  journal={arXiv preprint arXiv:2507.06261},
  year={2025}
}

@article{jaech2024openai,
  title={Openai o1 system card},
  author={Jaech, Aaron and Kalai, Adam and Lerer, Adam and Richardson, Adam and El-Kishky, Ahmed and Low, Aiden and Helyar, Alec and Madry, Aleksander and Beutel, Alex and Carney, Alex and others},
  journal={arXiv preprint arXiv:2412.16720},
  year={2024}
}

@article{wang2025nemotron,
  title={Nemotron-cascade: Scaling cascaded reinforcement learning for general-purpose reasoning models},
  author={Wang, Boxin and Lee, Chankyu and Lee, Nayeon and Lin, Sheng-Chieh and Dai, Wenliang and Chen, Yang and Chen, Yangyi and Yang, Zhuolin and Liu, Zihan and Shoeybi, Mohammad and others},
  journal={arXiv preprint arXiv:2512.13607},
  year={2025}
}

@inproceedings{schulman2015trust,
  title={Trust region policy optimization},
  author={Schulman, John and Levine, Sergey and Abbeel, Pieter and Jordan, Michael and Moritz, Philipp},
  booktitle={International conference on machine learning},
  pages={1889--1897},
  year={2015},
  organization={PMLR}
}

@inproceedings{madan2023learning,
  title={Learning gflownets from partial episodes for improved convergence and stability},
  author={Madan, Kanika and Rector-Brooks, Jarrid and Korablyov, Maksym and Bengio, Emmanuel and Jain, Moksh and Nica, Andrei Cristian and Bosc, Tom and Bengio, Yoshua and Malkin, Nikolay},
  booktitle={International Conference on Machine Learning},
  pages={23467--23483},
  year={2023},
  organization={PMLR}
}

@inproceedings{huang2025math,
  title={MATH-Perturb: Benchmarking LLMs’ Math Reasoning Abilities against Hard Perturbations},
  author={Huang, Kaixuan and Guo, Jiacheng and Li, Zihao and Ji, Xiang and Ge, Jiawei and Li, Wenzhe and Guo, Yingqing and Cai, Tianle and Yuan, Hui and Wang, Runzhe and others},
  booktitle={International Conference on Machine Learning},
  pages={25311--25328},
  year={2025},
  organization={PMLR}
}
\bibliographystyle{icml2026}


\newpage
\appendix
\onecolumn
\section{Derivations}
\label{app:proofs}
\subsection{Proof of Proposition \ref{prop:main}}
For a fixed question $\mathbf{x}$, and rollouts from $\pi_{\mathrm{old}}(\cdot \mid \mathbf{x})$, the expected loss is
\begin{align}
\mathcal{L}(\mathbf{x})
:=
\mathbb{E}_{\mathbf{y}\sim \pi_{\mathrm{old}}(\cdot\mid \mathbf{x})}
\left[
\left(
\log\!
        \frac{
            Z_{\phi}(\mathbf{x})\, \pi_{\theta}(\mathbf{y} \mid \mathbf{x})
        }{
            \pi_{\mathrm{old}}(\mathbf{y} \mid \mathbf{x})\,
            \exp\!\bigl(\beta^{-1} r(\mathbf{x}, \mathbf{y})\bigr)
        }
    \right)^2
\right].
\end{align}

By differentiating w.r.t.\ $\log Z_{\phi}(\mathbf{x})$, we have
\begin{align}
\frac{\partial \mathcal{L}(\mathbf{x})}{\partial \log Z_{\phi}(\mathbf{x})}
&=
\mathbb{E}_{\mathbf{y}\sim \pi_{\mathrm{old}}(\cdot\mid \mathbf{x})}
\left[
2\left(
\log Z_{\phi}(\mathbf{x})
+ \log \pi_{\theta}(\mathbf{y}\mid \mathbf{x})
- \log \pi_{\mathrm{old}}(\mathbf{y}\mid \mathbf{x})
- \beta^{-1} r(\mathbf{x},\mathbf{y})
\right)
\right].
\end{align}

Setting the derivative to zero yields
\begin{align}
\mathbb{E}_{\mathbf{y} \sim \pi_{\mathrm{old}}(\cdot\mid \mathbf{x})}
\left[
\log Z^{*}(\mathbf{x})
+ \log \pi_{\theta}(\mathbf{y}\mid \mathbf{x})
- \log \pi_{\mathrm{old}}(\mathbf{y}\mid \mathbf{x})
- \beta^{-1} r(\mathbf{x},\mathbf{y})
\right]
&=
0
,
\end{align}
hence
\begin{align}
\log Z^{*}(\mathbf{x})
&=
\mathbb{E}_{\mathbf{y}\sim \pi_{\mathrm{old}}(\cdot\mid \mathbf{x})}
\left[
\beta^{-1} r(\mathbf{x},\mathbf{y})
+ \log \pi_{\mathrm{old}}(\mathbf{y}\mid \mathbf{x})
- \log \pi_{\theta}(\mathbf{y}\mid \mathbf{x})
\right].
\end{align}

By rearranging, we have
\begin{align}
\mathbb{E}_{\mathbf{y}\sim \pi_{\mathrm{old}}(\cdot\mid \mathbf{x})}[r(\mathbf{x},\mathbf{y})]
&=
\beta \log Z^{*}(\mathbf{x})
-
\beta\,
\mathbb{E}_{\mathbf{y}\sim \pi_{\mathrm{old}}(\cdot\mid \mathbf{x})}
\!\left[
\log \pi_{\mathrm{old}}(\mathbf{y}\mid \mathbf{x})
-
\log \pi_{\theta}(\mathbf{y}\mid \mathbf{x})
\right]
\\
&=
\label{eq:before_sub}
\beta \log Z^{*}(\mathbf{x})
-
\beta
D_{\mathrm{KL}}\!\left(
\pi_{\mathrm{old}}(\cdot\mid \mathbf{x})
\;\big\|\;
\pi_{\theta}(\cdot\mid \mathbf{x})
\right).
\end{align}

Since reward is 0 for a wrong rollout and 1 for a correct rollout, we have
\begin{align}
\label{eq:E_is_p}
p_{\mathrm{old}}(\mathbf{x}) = \mathbb{E}_{\mathbf{y}\sim \pi_{\mathrm{old}}(\cdot\mid \mathbf{x})}[r(\mathbf{x},\mathbf{y})].
\end{align}

Substituting Eq. \ref{eq:E_is_p} to Eq. \ref{eq:before_sub} yields
\begin{align}
\label{eq:final}
p_{\mathrm{old}}(\mathbf{x})
&=
\beta \log Z^{*}(\mathbf{x})
-
\beta
D_{\mathrm{KL}}\!\left(
\pi_{\mathrm{old}}(\cdot\mid \mathbf{x})
\;\big\|\;
\pi_{\theta}(\cdot\mid \mathbf{x})
\right).
\end{align}

\subsection{Generalizing to Arbitrary Binary Reward Configurations}
\label{app:extension}
Rather than assigning a reward value of 0 to an incorrect output and 1 to a correct output, several prior works adopt alternative reward configurations. For example, DAPO \citep{yu2025dapo} assigns a reward of -1 to incorrect outputs and a reward of 1 to correct outputs. We show that the relationship between the partition function and accuracy estimates extends naturally to arbitrary reward configurations.

Assume that a reward value of $a$ is assigned to an incorrect output and a reward value of $b$ is assigned to a correct output, where $b\neq a$. Then, we can express $\mathbb{E}_{\mathbf{y}\sim \pi_{\mathrm{old}}(\cdot\mid \mathbf{x})}[r(\mathbf{x},\mathbf{y})]$ as follows:
\begin{align}
\mathbb{E}_{\mathbf{y}\sim \pi_{\mathrm{old}}(\cdot\mid \mathbf{x})}[r(\mathbf{x},\mathbf{y})]
&=
a\,(1-p_{\mathrm{old}}(\mathbf{x})) + b\,p_{\mathrm{old}}(\mathbf{x})
\\
&=
a + (b-a)\,p_{\mathrm{old}}(\mathbf{x}).
\label{eq:E_is_p_ab}
\end{align}
Expressing in terms of $p_{\mathrm{old}}(\mathbf{x})$ gives
\begin{align}
p_{\mathrm{old}}(\mathbf{x})
&=
\frac{
\mathbb{E}_{\mathbf{y}\sim \pi_{\mathrm{old}}(\cdot\mid \mathbf{x})}[r(\mathbf{x},\mathbf{y})] - a
}{b-a}.
\label{eq:p_from_Er_ab}
\end{align}
Substituting Eq.~\eqref{eq:before_sub} into Eq.~\eqref{eq:p_from_Er_ab} yields
\begin{align}
p_{\mathrm{old}}(\mathbf{x})
&=
\frac{
\beta \log Z^{*}(\mathbf{x})
-
\beta
D_{\mathrm{KL}}\!\left(
\pi_{\mathrm{old}}(\cdot\mid \mathbf{x})
\;\big\|\;
\pi_{\theta}(\cdot\mid \mathbf{x})
\right)
- a
}{b-a}.
\end{align}
Note that setting $a=0$ and $b=1$ recovers the result obtained in Eq. \ref{eq:final}.


\section{Experimental Details}
\label{app:experimental_details}
\subsection{Additional Training Details}
For training on Qwen2.5-Math-1.5B and Deepseek-Distill-Qwen-1.5B models, we use a single node consisting of 2 A6000 GPUs. For training on the Qwen2.5-Math-7B model, we use a single node consisting of 4 A100 GPUs. We use the verl \citep{sheng2025hybridflow} library for training and evalution across all runs. For all runs in the mathematical reasoning domain, we use a training budget of 150 steps. For runs on the code generation task, due to the substantial time required for training, we limit the training to 40 hours. As for the training and evaluation data, we follow the preprocessing pipeline from the official FlowRL repository\footnote{https://github.com/Xuekai-Zhu/FlowRL}. Other training hyperparameters kept constant across all runs are listed in Table \ref{tab:hyperparam}.

\begin{table}[t]
\caption{Hyperparameters used for training on the DeepScaleR and DeepCoder datasets. Unspecified hyperparameters inherit the default configurations under the verl library.}
\centering
\small
\begin{tabular}{lcc}
\toprule
\textbf{Hyperparameter} & \textbf{DeepScaleR} & \textbf{DeepCoder} \\
\midrule
Learning Rate (LR) & $1 \times 10^{-6}$ & $1 \times 10^{-6}$ \\
Gradient Clip &1.0 &1.0 \\
Optimizer & AdamW & AdamW \\
Weight Decay & 0.1 & 0.1 \\
Warm-up Steps & 10 & 10 \\
Global Batch Size & 128 & 64 \\
PPO Mini-batch Size & 32 & 32 \\
Micro-batch Size (per GPU) & 8 & 8 \\
Rollouts per Question ($N$) & 8 & 8 \\
Clip Ratio & 0.2 & 0.2 \\
KL Divergence Penalty & 0.0 & 0.0 \\
Entropy Coefficient & 0.0 & 0.0 \\
Rollout Temperature & 1.0 & 1.0 \\
Max Input Tokens & 1024 & 1024 \\
Max Response Tokens & 3072 & 3072 \\
\bottomrule
\end{tabular}
\label{tab:hyperparam}
\end{table}

\subsection{PACED-RL Implementation Details}
For the partition function $Z_{\phi}(\mathbf{x})$, following \citep{lee2025learning,zhu2026flowrl}, we use a 3-layer MLP on top of LLM last hidden layer embeddings. We use the last hidden layer embeddings of the frozen base model as the input to the $Z_{\phi}(\mathbf{x})$. The $Z_{\phi}(\mathbf{x})$ module is optimized using a seperate PyTorch optimizer, with the learning rate set to 1e-4. Lastly, after obtaining the accuracy estimates from Eq. \ref{eq:accuracy_final}, we clip the value to [0,1]. 

\begin{algorithm}[tb]
  \caption{PACED-RL}
  \label{alg:our_method}
  \begin{algorithmic}[1]
    \REQUIRE dataset $\mathcal{D}$; steps $T$; batch size $m$; rollout size $N$;
KL coefficient $\beta$; target accuracy $\tau$; buffer capacity $B_{\mathrm{max}}$; replay add count $B_{\mathrm{add}}$
    \STATE Initialize policy $\pi_{0}$ and partition network $Z_{\phi}$
    \STATE Initialize replay buffer $\mathcal{B}\leftarrow\emptyset$ with capacity $B_{\mathrm{max}}$
    \STATE Set $\pi_{\mathrm{old}} \leftarrow \pi_{0}$
    \FOR{$t=0$ {\bfseries to} $T-1$}
      \STATE Estimate accuracies:
      $\hat{p}_{\mathrm{old}}(\mathbf{x}) \leftarrow \beta \log Z_{\phi}(\mathbf{x}) \quad \forall\,\mathbf{x}\in\mathcal{D}$
      \STATE Select $m$ questions from accuracy estimates:
      $\mathcal{D}_{t}
\leftarrow
\operatorname*{argmin}_{\substack{\mathcal{D}_t \subseteq \mathcal{D} \\ |\mathcal{D}_t| = m}}
\sum_{\mathbf{x}\in \mathcal{D}_t}
\lvert\hat{p}_{\mathrm{old}}(\mathbf{x}) - \tau\rvert$
      \STATE Generate trajectories using $\pi_{\mathrm{old}}$:
      $\mathcal{T}_{t} \leftarrow
      \{(\mathbf{x}_i,\{\mathbf{y}_{i,j}\}_{j=1}^{N})\}_{i=1}^{m},
      \quad \mathbf{y}_{i,j}\sim \pi_{\mathrm{old}}(\cdot\mid \mathbf{x}_i)$
      \STATE Augment training batch with replay:
      $\widetilde{\mathcal{T}_{t}}\leftarrow \mathcal{T}_{t}\cup \mathcal{B}$
      \STATE Update $\pi_{t}$ to $\pi_{t+1}$, update $Z_{\phi}$ on $\widetilde{\mathcal{T}}_{t}$ using Eq.~\eqref{eq:our_tb_loss}
      \STATE Set $\pi_{\mathrm{old}} \leftarrow \pi_{t+1}$
      \STATE Filter out incorrect trajectories from $\mathcal{T}_{t}$:
$
\mathcal{T}_{t}' \;\leftarrow\;
\{(\mathbf{x}, \{\mathbf{y}_j\}) \in \mathcal{T}_{t}
\;\mid\;
 r(\mathbf{x}, \mathbf{y}_j) = 1 \}
$
      \STATE Select $B_{\mathrm{add}}$ samples from $\mathcal{T}_{t}^{'}$ to add to buffer:
        $
    \mathcal{C}_{t}
    \;\leftarrow\;
    \operatorname*{argmax}_{\substack{\mathcal{C}_t\subseteq \mathcal{T}_{t}^{'}, |\mathcal{C}_t|=B_{\mathrm{add}}}}
    \;\sum_{(\mathbf{x},\{\mathbf{y}_j\})\in \mathcal{C}_t}
    \Bigl|p_{\mathrm{old}}(\mathbf{x})-\hat{p}_{\mathrm{old}}(\mathbf{x})\Bigr|
    $
    \STATE Add replay candidates to buffer:
    $\mathcal{B}\leftarrow \operatorname{Push}(\mathcal{B},~\mathcal{C}_{t})$
    \ENDFOR
  \end{algorithmic}
\end{algorithm}
\subsection{Baseline Implementation Details}
For GRPO  \citep{shao2024deepseekmath} and DS \citep{yu2025dapo}, we use the implementation provided in the verl \citep{sheng2025hybridflow} library. For FlowRL \citep{zhu2026flowrl} and MoPPS \citep{qu2026can}, we adopt the official implementations and default hyperparameters provided by the authors in their implementations. For LILO \citep{foster2026lilo}, we implement the over-sampling and rejection-sampling procedure that selects the top-$m$ question prompts whose accuracies are closest to $0.5$. Following the original specification, at each training step we uniformly sample $4m$ prompts from the dataset and generate $N$ rollouts for each prompt. The $m$ prompts whose empirical accuracies are closest to $0.5$ are then selected for training, and the corresponding rollouts are reused for policy optimization.
\begin{figure*}[ht]
  \centering
  \includegraphics[width=0.7\textwidth]{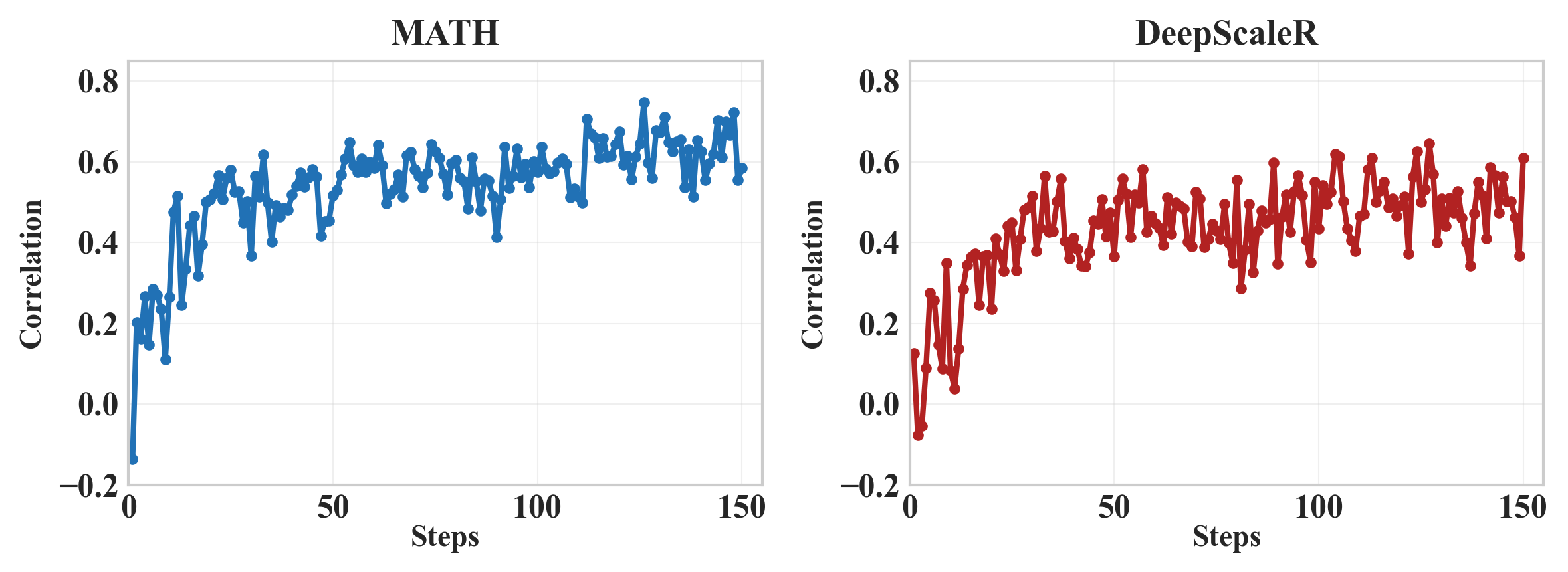}
  \caption{
    Spearman's correlation coefficient measured on the MATH training data and the DeepScaleR training data, using the Qwen2.5-Math-1.5B model.
  }
  \label{fig:spearmans}
\end{figure*}
\begin{figure*}[ht]
  \centering
  \includegraphics[width=0.7\textwidth]{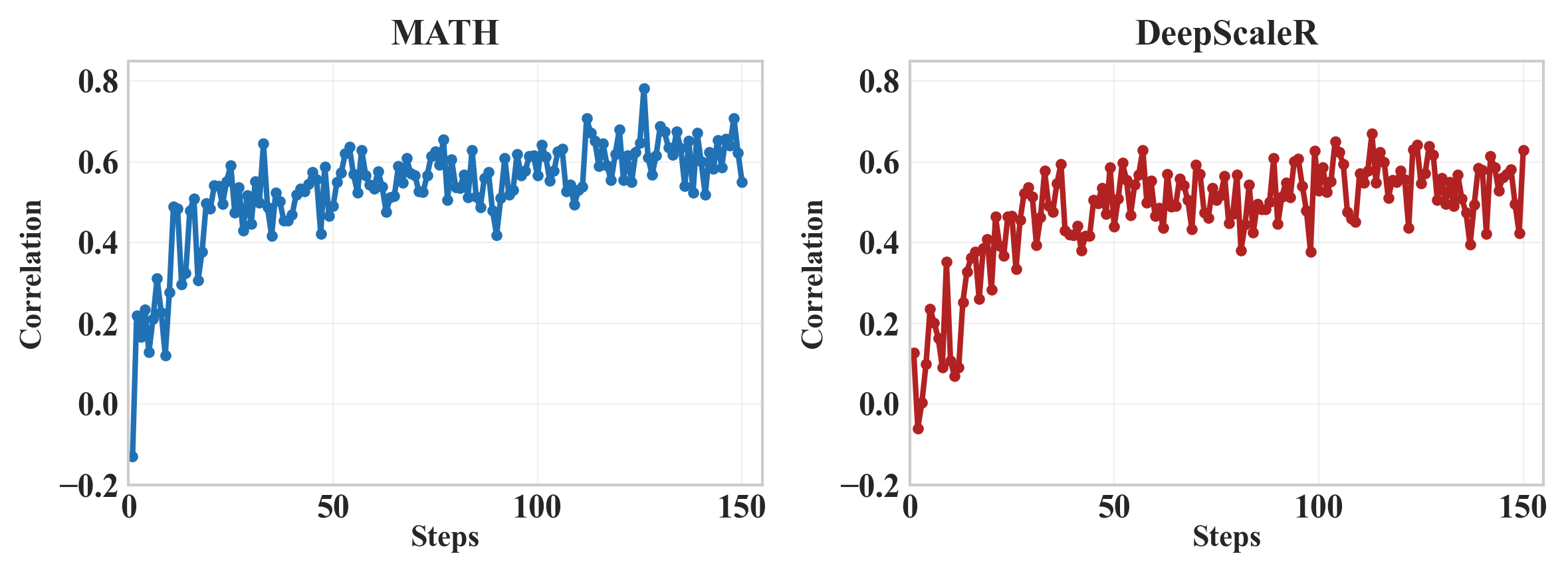}
  \caption{
    Pearson's correlation coefficient measured on the MATH training data and the DeepScaleR training data, using the Qwen2.5-Math-1.5B model..
  }
  \label{fig:pearsons}
\end{figure*}
\section{Supplementary Results}
\label{app:supplementary_results}
\subsection{Correlation of Accuracy Estimates to Empirically Observed Accuracies}
Fig. \ref{fig:spearmans} and Fig. \ref{fig:pearsons} show the Spearman's correlation coefficient and Pearson's correlation coefficient for randomly sampled question prompts over the training process, on the Qwen2.5-Math-1.5B model. The constantly high $(> 0.5)$ correlation values after $\sim20$ training steps indicate that the online accuracy estimates from $Z_{\phi}$ serve as reliable estimators.
\subsection{Robustness of the Accuracy Estimator to Adversarial Inputs}
Due to the light-weighted nature of the accuracy estimator, the accuracy estimation may be sensitive to adversarial cases in which small textual perturbations induce large changes in question difficulty. To examine this failure mode, we use the MATH-Perturb benchmark \citep{huang2025math}, which constructs two perturbed variants for each original question: \textit{Simple}, where small edits preserve the solution structure and difficulty, and \textit{Hard}, where small but semantically important edits change the solution and increase the difficulty. An illustrative example of a sample from the Math-Perturb benchmark is as follows:

\begin{quote}
\textbf{Original question:} Find the range of
\[
y = \frac{x^2 + 3x + 2}{x + 1}.
\]

\textbf{Simple variant:} Find the range of
\[
y = \frac{x^2 + 3x + 2}{x + 2}.
\]

\textbf{Hard variant:} Find the range of
\[
y = \frac{x^2 + 3x + 2}{x}.
\]
\end{quote}

Since the original questions are not explicitly released, we compare the \textit{Simple} and \textit{Hard} variants directly. Specifically, throughout training, we measure the pairwise accuracy of the estimator in Eq.~\ref{eq:accuracy_final}, defined as the fraction of pairs for which the \textit{Hard} variant is assigned a lower predicted accuracy than the corresponding \textit{Simple} variant. As shown in Table~\ref{tab:pairwise_accuracy}, results with Qwen2.5-Math-1.5B show that the estimator correctly identifies the \textit{Simple} variant as easier in the majority of cases throughout training, even under this adversarial perturbation setting. Taken together with the high correlations shown in Fig. \ref{fig:spearmans} and Fig. \ref{fig:pearsons}, we view our estimator as a useful low-cost proxy for difficulty, sufficient for an adaptive learning strategy. We leave further studies on improving sensitivity to prompt perturbations for future works.
\begin{table}[ht]
\caption{Pairwise accuracy over the course of training.}
\centering
\setlength{\tabcolsep}{3pt}
\begin{tabular}{lc}
\toprule
\textbf{Step} & \textbf{Pairwise Accuracy (\%)} \\
\midrule
30  & 65.2 \\
60  & 67.0 \\
90  & 67.7 \\
120 & 69.9 \\
150 & 69.5 \\
\bottomrule
\end{tabular}
\vspace{3pt}
\label{tab:pairwise_accuracy}
\end{table}
\section{Limitations}
\label{app:limitations}
While we evaluate PACED-RL on models with 1.5B and 7B parameters and observe consistent improvements across both scales, we do not explore substantially larger models due to computational constraints. As a result, it remains an open question whether the gains of PACED-RL persist when applied to substantially larger LLMs.

In addition, although our experiments span both mathematical reasoning and code generation tasks, we focus exclusively on domains where verifiable reward signals are available. We do not study non-verifiable settings such as preference optimization or open-ended creative generation. Extending PACED-RL to these domains may require more sophisticated reward modeling or alternative accuracy estimation mechanisms, which we leave to future work. 

Lastly, our low-cost accuracy estimation based on the learned partition function $Z_{\phi}$ relies on the assumption that the term
$
\beta\, D_{\mathrm{KL}}\!\left(\pi_{\mathrm{old}}(\cdot \mid \mathbf{x}) \,\big\|\, \pi_{\theta}(\cdot \mid \mathbf{x})\right)
$
remains sufficiently small during training and can therefore be safely neglected. While our empirical results demonstrate that PACED-RL yields consistent performance improvements in both code generation and mathematical reasoning domains, the validity of this assumption may depend on the characteristics of the task and training dynamics. In particular, when extending PACED-RL to other domains, this approximation should be carefully examined, as deviations from this regime could degrade the accuracy of the estimator. We leave a systematic investigation of this assumption across broader domains to future work.
\end{document}